%% file: main.tex
\documentclass[twoside]{article}

\usepackage[accepted]{aistats2021}

\usepackage{microtype}
\usepackage{graphicx}
\usepackage{subfigure}
\usepackage{booktabs} 
\usepackage{float}
\usepackage{amsfonts}
\usepackage{multirow}
\usepackage[ruled,vlined]{algorithm2e}

\usepackage{hyperref}
\usepackage{enumitem}

\usepackage[round]{natbib}
\renewcommand{\bibname}{References}
\renewcommand{\bibsection}{\subsubsection*{\bibname}}

\begin{document}

\runningauthor{Zhang, Rajan, Pineda, Lambert, Biedenkapp, Chua, Hutter, Calandra}

\twocolumn[

\aistatstitle{On the Importance of Hyperparameter Optimization for Model-based Reinforcement Learning}

\vspace*{-15pt}
\aistatsauthor{
  Baohe Zhang\\
  University of Freiburg
   \vspace*{-5pt}
  \And
  Raghu Rajan\\
  University of Freiburg
   \vspace*{-5pt}
  \And
  Luis Pineda\\
  Facebook AI Research
   \vspace*{-5pt}
  \AND
  Nathan Lambert\\
  UC Berkeley
   \vspace*{-5pt}
  \And
  Andr\'e Biedenkapp\\
  University of Freiburg
   \vspace*{-5pt}
  \And
  Kurtland Chua\\
  Princeton University
   \vspace*{-5pt}
  \AND
  Frank Hutter\\
  University of Freiburg and Bosch Center for AI
  \And
  Roberto Calandra\\
  Facebook AI Research
   \vspace*{35pt}
}
]

\begin{abstract}
\input{0_abstract.tex}
\end{abstract}

\section{Introduction}
\label{sec:intro}
\input{1_introduction.tex}

\section{Related Work}
\label{sec:related}
\input{2_related.tex}

\section{Background}
\label{sec:backgrounds}
\input{3_backgrounds.tex}

\section{HPO for MBRL}
\label{sec:approach}
\input{4_approach.tex}

\section{Experimental Evaluation}
\label{sec:results}
\input{5_results.tex}

\section{Conclusion}
\label{sec:conclusion}
\input{6_conclusion.tex}

\clearpage

\subsubsection*{Acknowledgements}
\label{sec:acknowledgement}
\input{7_acknowledgement.tex}

\bibliography{references}

\clearpage

\onecolumn
\appendix
\section{SUPPLEMENTARY MATERIAL}
\label{sec:appendix}
\input{8_appendix.tex}

\end{document}

%% file: 0_abstract.tex
Model-based Reinforcement Learning (MBRL) is a promising framework for learning control in a data-efficient manner. 
MBRL algorithms can be fairly complex due to the separate dynamics modeling and the subsequent planning algorithm, and as a result, they often possess tens of hyperparameters and architectural choices. 
For this reason, MBRL typically requires significant human expertise before it can be applied to new problems and domains. 
To alleviate this problem, we propose to use automatic hyperparameter optimization (HPO). 
We demonstrate that this problem can be tackled effectively with automated HPO, which we demonstrate to yield significantly improved performance compared to human experts.
In addition, we show that tuning of several MBRL hyperparameters dynamically, i.e. during the training itself, further improves the performance compared to using static hyperparameters which are kept fixed for the whole training.
Finally, our experiments provide valuable insights into the effects of several hyperparameters, such as \emph{plan horizon} or \emph{learning rate} and their influence on the stability of training and resulting rewards.

%% file: 1_introduction.tex
Model-based Reinforcement Learning~(MBRL) is a powerful framework for learning to solve continuous control tasks.
Recent work in MBRL has demonstrated performance comparable to traditional model-free algorithms, using only a fraction of the data~\citep{Williams2017,chua2018deep,DBLP:conf/iclr/KurutachCDTA18,DBLP:conf/nips/AmosRSBK18}.
Modern MBRL algorithms are complex tools which rely on several sub-components: the training of a dynamics model, the prediction and propagation of state-action trajectories, and the nested optimization of a policy or the planning of the action sequence.
Each of these sub-components, in turn, requires the tuning of several design parameters that can significantly impact performance. 
As a result, MBRL algorithms require a high degree of human expertise to configure before they can be applied to new tasks.

The observation that hyperparameter tuning can be a tedious and domain-specific task which requires high levels of expertise and computational cost is not unique to MBRL. 
Hyperparameter Optimization~(HPO) emerged as a promising and quickly growing approach to automate the training of machine learning models \citep[see e.g. the survey by][]{feurer-automl19}.
HPO has been successfully applied to several domains such as natural language processing~\citep{DBLP:conf/iclr/MelisDB18} or deep RL \citep{henderson2018deep} to improve the performance and robustness of trained models which are highly dependent on their hyperparameter settings.
Although HPO has been previously used in conjunction with model-free RL algorithms~\citep{jaderberg2017population, henderson2018deep, DBLP:conf/nips/PaulKW19},
to the best of our knowledge, HPO has not yet been evaluated in the MBRL setting.
Using HPO with MBRL poses challenges as MBRL is often used in the low-data regime, which constrains HPO to be as data-efficient as possible, and MBRL typically has a larger number of hyperparameters that have unknown interaction effects between each other. Additionally, MBRL provides another exciting challenge to HPO algorithms through its potential non-stationarity, where during the learning process different hyperparameter settings are required to improve learning quality.
Commonly, HPO assumes a stationary process, where a single hyperparameter setting works best during the whole learning process.
In this paper, we propose to utilize (dynamic) HPO for MBRL to drastically reduce the need for human expertise and further improve performance.

Specifically, we demonstrate: $1)$ \textit{automatic} HPO can outperform human experts in tuning MBRL algorithms, achieving new state-of-the-art performance (to the point of training in one experiment an agent so successful that it ``breaks down'' the simulator); $2)$ \textit{dynamic} HPO can further improve the performance compared to \textit{static} HPO; $3)$ the best hyperparameter configurations differ across MBRL scenarios; $4)$ the dramatic effect that different hyperparameter choices have in MBRL.

By demonstrating the value that HPO can bring to MBRL approaches, we envision a future where HPO can be widely adopted to drive further progress in the MBRL community, while at the same time creating a new research direction and challenges to stimulate the HPO community.

%% file: 2_related.tex
\paragraph{HPO}
Random search and grid search are simple and commonly used methods for performing HPO~\citep{bergstra-jmlr12}. 
A more informed framework for HPO is Bayesian Optimisation~\citep[BO;][]{snoek2012practical,bergstra2011algorithms,springenberg2016bayesian,hutter2011sequential}. 
BO sequentially optimizes an algorithm's hyperparameters by using previous evaluation results to select the next set of hyperparameters to evaluate.
The previous evaluations are used to build a surrogate which models the relationship between hyperparameter values and resulting performance.
Then, an acquisition function is used to trade off exploration and exploitation.
The sequential nature of BO and use of full function evaluations however can make it very costly.
Multi-fidelity HPO~\citep{DBLP:journals/jair/KandasamyDOSP19} speeds up HPO by using evaluations of an algorithm on a cheaper, low fidelity to judge how an algorithm would perform when evaluated on an expensive, high fidelity. 
For example \citet{klein-aistats17} used the number of datapoints as fidelity, when tuning the hyperparameters of an SVM.
On smaller subsets of the data many more hyperparameter configurations can be evaluated, compared to evaluating only a few configurations on the full data  for the same cost, given that performance on the small dataset is correlated with performance on the full dataset.
Successive Halving~\citep[SH;][]{jamieson2016non} used random search to sample a set of configurations to be run on the lowest fidelity and only the best performing configurations continue to be optimized further on higher ones.
This allows SH to speed up the search for well performing hyperparameters. 
However, it is not always clear what minimum budgets would serve as good fidelities for the full budget. 
To overcome this, Hyperband~\citep{li2017hyperband} launches multiple Successive Halving runs with different minimum budgets. 
BOHB \citep{DBLP:conf/icml/FalknerKH18} further builds on Hyperband and BO by using a surrogate model to guide Hyperband's search procedure.
The approaches discussed so far only search for \textit{static} hyperparameter settings. 
Population based training \citep[PBT;][]{jaderberg2017population}, on the other hand, was proposed to facilitate \textit{dynamic} optimization of non-stationary processes.
Instead of resulting in a static configuration, which is constant throughout a whole run of the optimizee, PBT changes the configuration at multiple stages.
Thus, PBT can adapt hyperparameters to a potentially changing search landscape.

\paragraph{HPO for model-free RL}
RL is known for its sensitivity to its hyperparameters as well as its neural architecture choices.
\citet{islam2017reproducibility} described the impact of hyperparameters and network architectures for policy gradient methods.
They showed, for example, the impact the activation can have on the achievable reward of TRPO and DDPG agents.
\citet{henderson2018deep} followed with a broader study, including more algorithms with a larger search space.
Multiple hyperparameter optimization methods have been used to optimize model-free reinforcement learning agents on a broad variety of tasks. 
\citet{DBLP:conf/icml/FalknerKH18} proposed to use BOHB to efficiently search for well performing hyperparameter settings and neural architectures of PPO.
Similarly, under the heading of AutoRL, \citet{runge2019learning} used BOHB for joint optimization of an RL agent's hyperparameters and neural architecture, whereas \citet{Chiang18AutoRL} proposed a two stage approach to optimize the agent hyperparameters and the network architectures.
The presented methods so far, however, ignored a crucial issue for RL:~the possible non-stationarity of the RL problem induces a possible non-stationarity of the hyperparameters. 
Thereby, at different stages of the learning process, different hyperparameter settings might be required to behave optimally.
For A3C-like model-free RL methods on various tasks, PBT dynamically optimized hyperparameters and achieved better performance over a random search baseline~\citep{jaderberg2017population}. 
\citet{DBLP:conf/nips/PaulKW19} proposed an alternative to PBT for policy gradient algorithms by utilizing the properties of policy gradient methods and learn a schedule for the hyperparameters via importance sampling. 

\paragraph{HPO for model-based RL}
To the best of our knowledge, we are the first to study hyperparameter optimization in the context of model-based reinforcement learning.
\citet{wang-arxiv19a} presented an extensive benchmarking study of different MBRL agents.
To compare the agents fairly, they only used static grid searches over the hyperparameters.
They report only the performance of the optimized agents and do not quantify the impact of the used grid-searches.

%% file: 3_backgrounds.tex
Among the HPO methods, we introduce PBT here in more detail because we provide an in-depth evaluation of novel settings for its training data usage and HPO methodology later.
\subsection{Population Based Training}\label{sec:PBT}
Population based training (PBT) is an evolutionary approach for dynamic HPO and allows to optimize hyperparameters during the training of the members of its population.
PBT starts out with a randomly initialized population. As a result, all members of its population start from different regions in the hyperparameter configuration space.
The members are ranked according to their current performance at regular intervals.
The worst performing members in the population are then replaced by the best ones, by copying over their parameters (network weights, in case of neural networks NNs)), as well as their hyperparameters in an \textit{exploitation step}. It is unclear from the original PBT paper, however, whether the data on which the members are trained are also copied over. We discuss ablations of this setting in the experiments section.
To allow searching for potentially better performing hyperparameter configurations, the copied hyperparameters are perturbed to allow for small steps in their immediate neighborhoods in an  \textit{exploration step}.
Over time, different configurations are evaluated \emph{during} the training process and potentially kept if they improve performance.
PBT comes with its own hyperparameters. In the original paper, members from the population are selected using \textit{truncation} during the exploitation step. Thereby, agents from the bottom $20\%$ of the population are replaced by agents from the top $20\%$. Additionally, continuous hyperparameter values are multiplied at random by $0.8$ or $1.2$ in the exploration step.

\subsection{Model-based Reinforcement Learning}\label{sec:MBRL}
We briefly introduce the reinforcement learning (RL) framework in the finite time horizon setting.
Let $(\mathcal{S}, \mathcal{A}, P, r)$ be a Markov Decision Process, where $\mathcal{S}$ is the set of states, $\mathcal{A}$ is the set of actions, $P$ represents transition dynamics, and $r$ is the reward function.
Furthermore, let $T$ denote a finite time horizon.
The objective is then to find a policy $\pi$ that maximizes the expected sum of rewards given by
\begin{equation}
\mathbb{E}\left[\sum_{t=0}^{T}r(s_t, a_t)\right],
    \label{eq:rl_obj}
\end{equation}
where $s_{t + 1} \sim P(\cdot | s_t, a_t)$ and $a_t \sim \pi(\cdot | s_t)$.

RL algorithms can be roughly divided into two categories, \emph{model-based} and \emph{model-free}, depending on whether they do or do not build a model of the environment dynamics, respectively. In the model-based approach, considered in this work, we first use data to learn a dynamics model, $\hat{P}$, approximating the true transition dynamics, $P$. Once this model is learned, a model-based method then optimizes the objective (Equation~\ref{eq:rl_obj}), using $\hat{P}$ in place of $P$. The resulting policy is then executed in the environment, typically for a limited number of steps to reduce the effect of compounding modeling errors, and the optimization/execution process is repeated in an inner iteration loop termed a \textit{trial}. The outer loop of learning and trials are interleaved for $n$ trials. Algorithm~\ref{alg:mbrl} summarizes the general MBRL framework.
For more information on MBRL we refer the reader to \citep{Deisenroth2011PILCO, chua2018deep}.

\begin{algorithm}[t]
        \KwIn{Number of trials, $n$, dataset, $D$, policy, $\pi$}
        \KwOut{Dynamics model $\hat{P}(s_{t+1}|s_t, a_t)$}
        Initialize dataset, $D$, from random actions \\
        Set trial counter, $i = 0$
        
        \While{$i < n$}{
            Train dynamics model $\hat{P}(s_{t+1}|s_t, a_t)$ on $D$\\
            \While{trial not done}{
                Optimize objective using $\hat{P}$ to obtain $\pi$ \\
                Execute $\pi$ and collect data $D'$ with controller using $P$ in real environment  \\
                Aggregate dataset $D = D \cup D'$ 
            }
        }
    \caption{Generic Model-based RL framework}
    \label{alg:mbrl}
\end{algorithm}

%% file: 4_approach.tex
Reinforcement learning involves challenging optimization problems due to the stochasticity of evaluation, high computational cost and possible non-stationarity of the hyperparameters.
Therefore, we now discuss some significant design decisions facing an RL practitioner.

\subsection{Comphrehensive Considerations}
\paragraph{Transferability across environments}
When transferability across \textit{environments} is the preferred goal of a set of hyperparameters, users expect them to be broadly applicable to related tasks.
For example, when training agents to play a variety of Atari games, hyperparameters are set such that the agent performs well on a variety of different games \citep{mnih-nature13}.
Alternatively, we could choose the performance on a specific task to be the main focus of hyperparameter tuning.
In the context of MBRL this could, for example, be motivated by the dynamics model which is specific to the environment at hand, often requiring very different hyperparameter settings to be learned as accurately and data-efficiently as possible \citep{wang-arxiv19a}.
Thus hyperparameter settings might not always transfer between environments.
Conversely, desirable hyperparameter settings might be those that give us the best performance on specific tasks.
As a consequence, when choosing between available HPO methods for MBRL, we need first to decide if we require robust configurations that transfer or if we require highly specific configurations that achieve the best possible result on a particular task.

\paragraph{Transferability across multiple runs}
The required robustness of the found configuration has consequences on the nature of the HPO method best suited for the current task.
Dynamic configurations can be expected to be less transferable across even \textit{multiple runs} than static ones.
By design, dynamic configurations make many more choices about the parameter settings than static configurations, thus making it very challenging to tune dynamic configurations by hand and without automatic HPO methods.
With an increasing number of decision points, it becomes more and more likely that each choice we make is specifically tailored to the environment and even the current run at hand.

\subsection{Design Choices}
\paragraph{Considered Methods}
Based on the mentioned considerations, we decided to investigate three approaches for the hyperparameter tuning of MBRL agents. 
Namely, we use Hyperband as a multi-fidelity approach; PBT as a dynamic configuration method; and random search as a baseline in HPO. As mentioned in Section~\ref{sec:backgrounds}, for PBT, we additionally evaluate the effect of copying not only the model but also the training data when members are exploited, such that the dynamics model is not learned from drastically different data.
We further consider a variation of PBT which we dub \emph{PBT with backtracking~(PBT-BT)}, to gain further insights into dynamic configuration. 
PBT-BT is similar to the work by \citet{dasagi2019ctrlz}, which allows agents to backtrack to checkpoints before performance dropped significantly, with the exception that PBT-BT not only backtracks the model parameters but also the hyperparameters.
We expect this modification to reduce the typical performance drops seen for RL training curves and to better explore unexplored parts of the hyperparameter search space from previously seen good configurations. For implementation details and hyperparameter settings of the used HPO methods, we refer to Appendix~\ref{appendix:pbt-bt}.

\paragraph{Handling Stochasticity of Evaluation}
In the context of MBRL, it is possible to optimize the hyperparameters based on the model loss, which is usually the Mean Squared Error (or likelihood for probabilistic models) of one-step predictions from fitting the model.
However, \citet{lambert2020objective} showed that the correlation between model fitting and final returns can be weak.
Alternatively when tuning the hyperparameters we can use the average returns over $n$ steps as HPO objective.
If $n$ is too small we might be prone to accept highly volatile hyperparameter settings as well performing, whereas if $n$ is too large we might make little progress in identifying truly well-performing configurations.

\paragraph{Handling Computational Demands}
As discussed in Section~\ref{sec:related}, traditional HPO methods such as BO can be computationally demanding.
Thus, when dealing with expensive MBRL problems we need to decide how we can limit the additional computational overhead required for HPO.
From the HPO literature we know that multi-fidelity approaches~\citep{li2017hyperband} as well as PBT allow for efficient parallelization and provide significant improvements.

%% file: 5_results.tex
\begin{figure*}[t]
\centering
\includegraphics[width=0.9\linewidth]{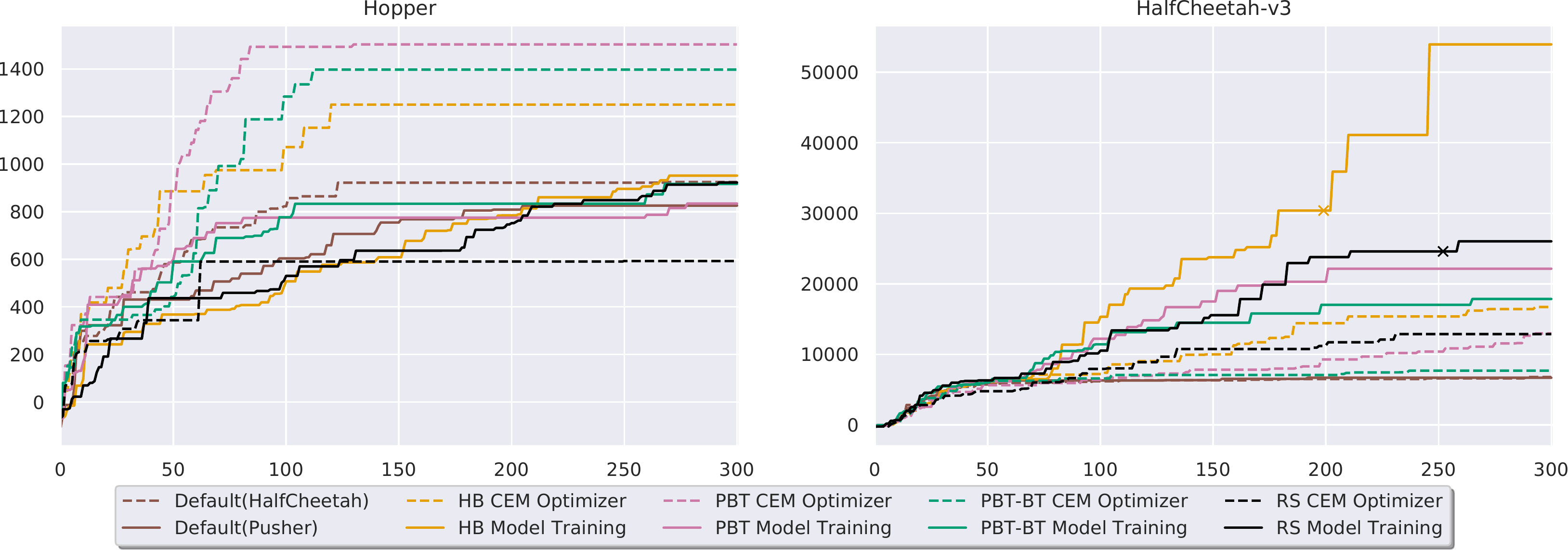}
\caption{Evaluation of the \emph{best} found hyperparameters and schedules from all search methods. Scores~(y-axis) are the maximum of the average return over $5$ evaluations over number of trials~(x-axis). Brown lines are the baselines using the hyperparameters tuned by a human expert. Markers in HalfCheetah show that the evaluation started to encounter a bug in Mujoco which caused numerical overflow. Hyperparameter schedules do not transfer across environments or even across runs on the same environment (for the latter, compare performances in the transfer experiments here with the actual HPO experiments in Figure \ref{fig:pbt_vs_rs_res}).}
\label{fig:eval_compare}
\end{figure*}
\label{results}
To address the importance of automatic HPO and validate our hypothesis of the non-stationarity of MBRL,
we compare automatic HPO results to hyperparameters tuned by human experts and analyze the performance gain of dynamic tuning over static tuning.
\subsection{Setting}
\label{exp_setting}

\paragraph{Optimizee}
To demonstrate the importance of HPO for MBRL we use the current state-of-the-art \emph{Probabilistic Ensembles With Trajectory Sampling~(PETS)}~\citep{chua2018deep} algorithm as optimizee.
PETS uses an ensemble of neural networks to learn a model of the environment which provides aleatoric and epistemic uncertainty estimates. 
In PETS, the dynamics model is chosen to be an ensemble of neural networks whose outputs parameterize anisotropic Gaussians.
Model Predictive Control~(MPC) is then used to get the policy, by directly optimizing the expected sum of rewards over a fixed planning horizon.
PETS, in particular, performs model predictive control with a cross-entropy method~(CEM) optimizer for action selection.
To evaluate an action sequence, PETS first samples a model from the ensemble, and rolls out the action sequence using the selected model, and computes the sum of rewards.
Action sequences are then evaluated by performing this process multiple times iteratively and averaging the simulated returns over the ensemble members.

\paragraph{Environments}
For the experiments, we consider four test environments: \emph{Pusher}, \emph{Reacher}, \emph{Hopper}, \emph{Halfcheetah} from MuJoCo~\citep{todorov2012mujoco} and a simulation environment of \emph{Daisy}, a robot hexapod to accomplish locomotion tasks. The reward signal for \emph{Daisy} is similar to Hopper and HalfCheetah, where we use the forward speed as the reward signal.
The number of trials is fixed to $80$ for pusher and $300$ for the rest, with rollout lengths of $150$ and $1\,000$ steps, respectively. 
The actions in the initial trial are sampled randomly to collect data to train the model before it is used by PETS in future trials.
We use Hopper and HalfCheetah for illustrative plots in the main paper; quantitative results for the other three tasks can be found in Appendix~\ref{appendix:results}.

\paragraph{Configuration Space}
We split the hyperparameters of PETS into two groups:
(i) \textit{Model Training} and (ii) \textit{CEM Optimizer}; to clearly differentiate the influence these parameter spaces have in the MBRL setting.
This allows optimizers to learn interaction effects within each group. 
When optimizing one group of hyperparameters, the others are set to the default value of the best manually tuned PETS hyperparameters as reported by \citet{chua2018deep}. We also optimized all hyperparameters together in Appendix~\ref{appendix:joint}. The full configuration spaces are given in Appendix~\ref{appendix:searchspaces}.

\paragraph{HPO Objective}
We use the average returns of the $3$ most recent trials as the objective for all HPO methods. 
This gives us a better estimate of the noisy reward in the MBRL tasks.
As discussed in Section~\ref{sec:approach}, we consider two scenarios: $1)$ the transferability of the hyperparameter schedule (static or dynamic) learned by the HPO methods across environments; and $2)$ where we are interested in the final learned model and policy reward across multiple runs. 
In the first scenario, we consider the mean performance of the top $5$ members during the search.
For the latter scenario, we use the best found schedules of each HPO method and report the performance of PETS over $5$ seeds.

\begin{figure*}[h]
    \centering
    \includegraphics[width=0.9\linewidth]{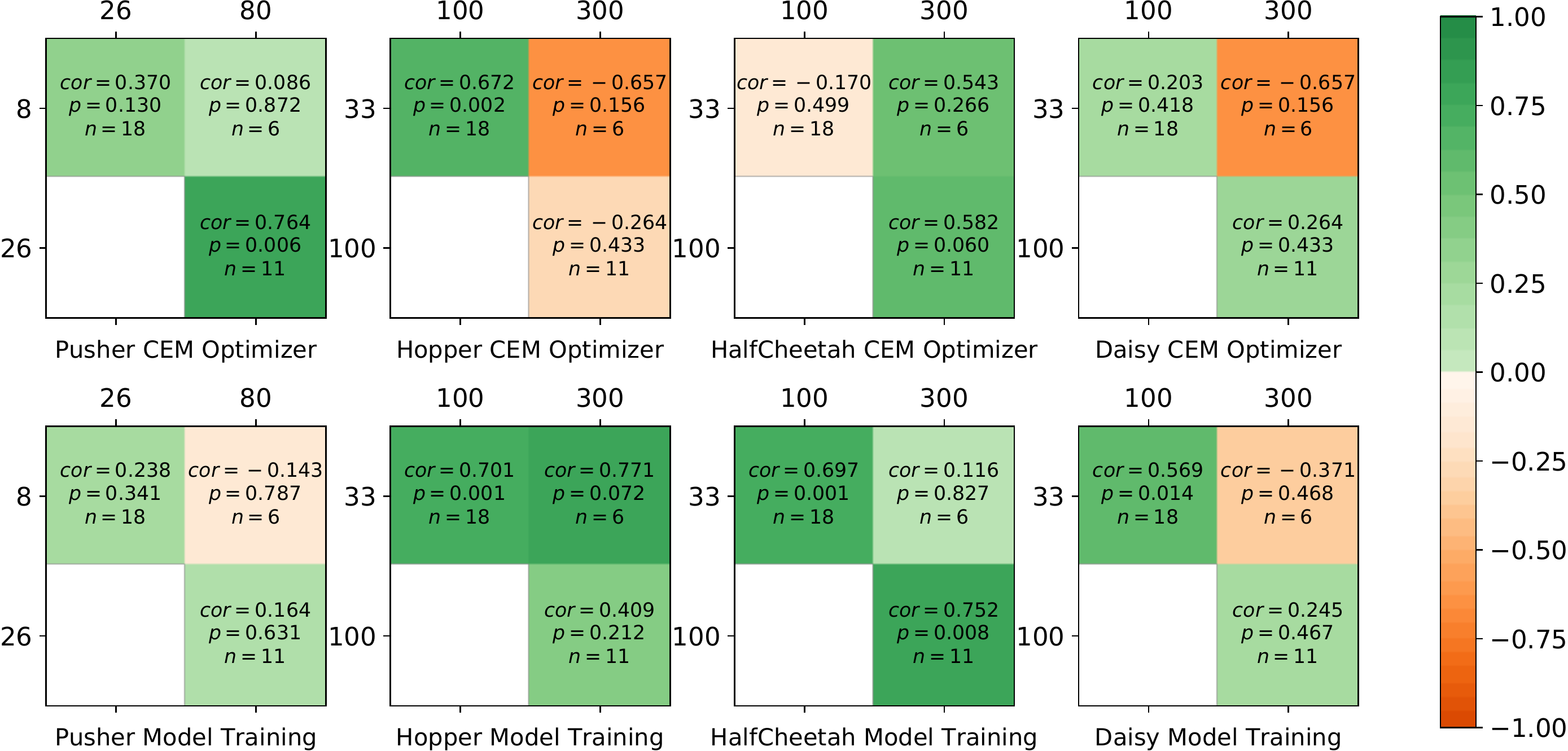}
    \caption{\label{fig:correlation} Spearman rank correlation of hyperparameters sampled by Hyperband across fidelities (i.e., number of trials) when training agents with $8$ and $80$ trials as the minimal/maximal budgets for Pusher and $33$ and $300$ trials for the rest of the environments.
    $cor$ is the correlation coefficient, $p$ is the p-value for the hypothesis test and $n$ gives the number of configurations trained in both fidelities. Weak or even negative correlation across different budgets are found in most of the environments, which shows different configurations perform best for different training durations and that dynamic tuning may be needed.}
\end{figure*}

\subsection{Importance of HPO}
To demonstrate the importance of HPO in the context of MBRL, we use the default hyperparameters tuned (on Pusher and on HalfCheetah) by a human  expert~\citep{chua2018deep} as a baseline.
We compare the performance of these baselines to the best hyperparameter settings found through means of each HPO method, see Figure~\ref{fig:eval_compare}.
If HPO was of low importance in MBRL, we would expect the hand-crafted hyperparameter settings to transfer well between environments and to perform competitively with the hyperparameter settings found by HPO.
Instead, we can observe that both default hyperparameter settings are outperformed by configurations obtained either through static or dynamic HPO.
In Figure~\ref{fig:eval_compare}, the Hopper and HalfCheetah environments show respectively a $1.5-2\times$ and $10\times$ improvement in final returns of HPO tuned agents over manually configured baselines which themselves involved substantial manual effort. The gains over manually tuned agents are not always so significant as can be seen in Appendix \ref{appendix:results}, especially for simpler environments, however, automatic tuning always outperforms manual tuning and saves significant overhead.

\subsection{Need for Dynamic Tuning}
We now use Hyperband to motivate the need for dynamic tuning. 
As fidelities in this setting we consider the number of trials used to train PETS.
We show that different configurations perform well across different fidelities in many instances. This reveals the possible need for dynamic tuning.
We calculate the Spearman correlation between the ranks of different configurations based on different fidelities of a Hyperband run (see Figure~\ref{fig:correlation}).
The correlations are very small or even negative, indicating that \emph{static} hyperparameter settings that are best for smaller fidelities do not perform as well or even worse on higher fidelities. 
Further we can observe for nearly all environments that configurations found on the lowest fidelity have a lower correlation to the highest fidelity than to the middle fidelity.
This suggests that different hyperparameter configurations work best for different training durations and that we may need to tune hyperparameters dynamically in order to perform better.
The outlier seems to be the model training parameters on hopper, which are highly correlated throughout. One possible reason is the impact of model training parameters on the performance is relatively low in Hopper, also shown in Figure~\ref{fig:eval_compare}.

\begin{figure*}[t]
\centering
\includegraphics[height=9cm]{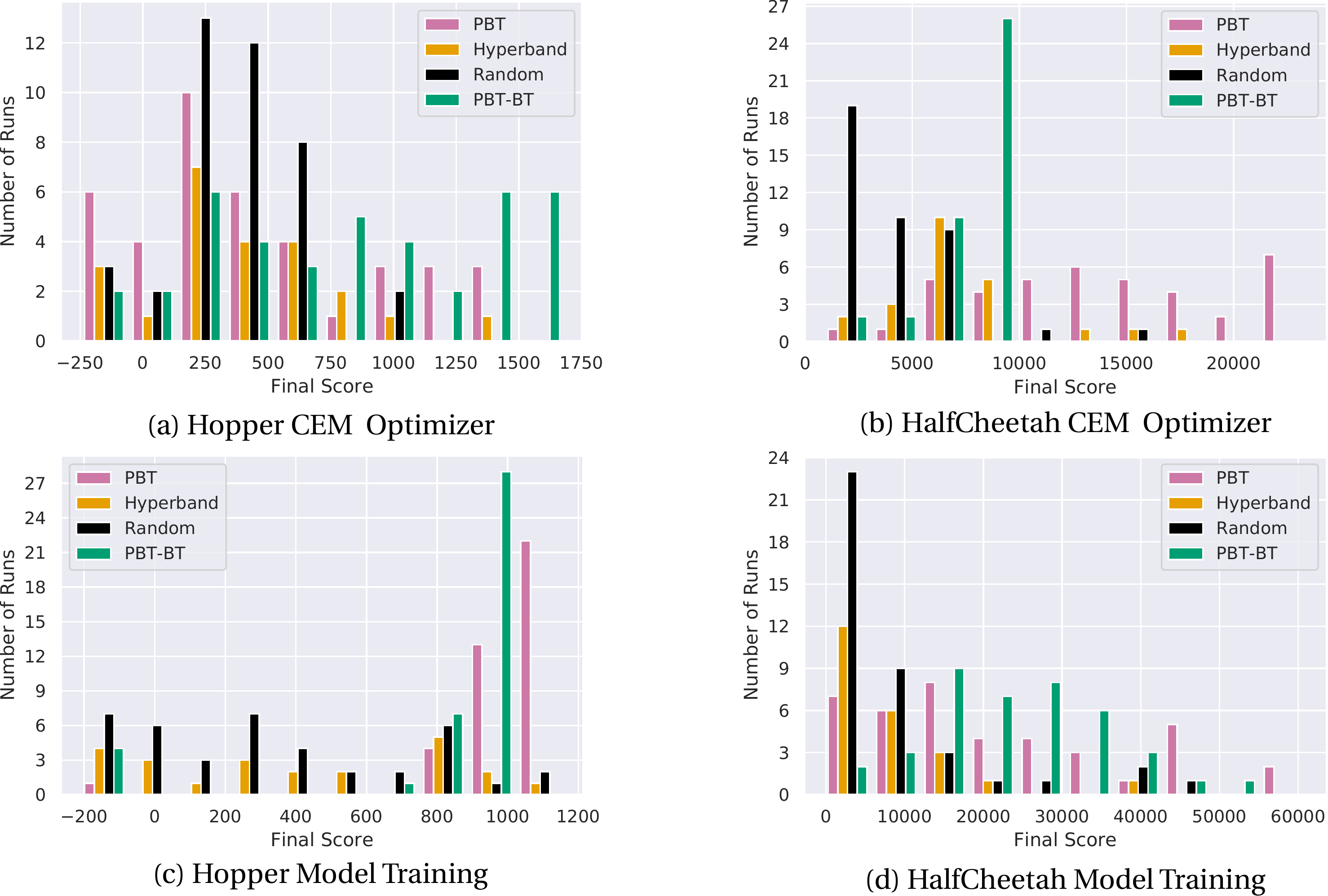}
 \caption{Histogram of achieved final episode return on Hopper and HalfCheetah over all configurations of CEM hyperparameters (top) and model training (bottom). We see that the importance of hyperparameters differs in different environments and that PBT can allocate more resources in well-performing areas during exploitation.
}
\label{fig:hist_all_searches}
\end{figure*}

\subsection{Breaking the Simulation}
\begin{figure*}[t]
\centering
    \subfigure[Hopper CEM Optimizer]{\includegraphics[width=0.43\textwidth]{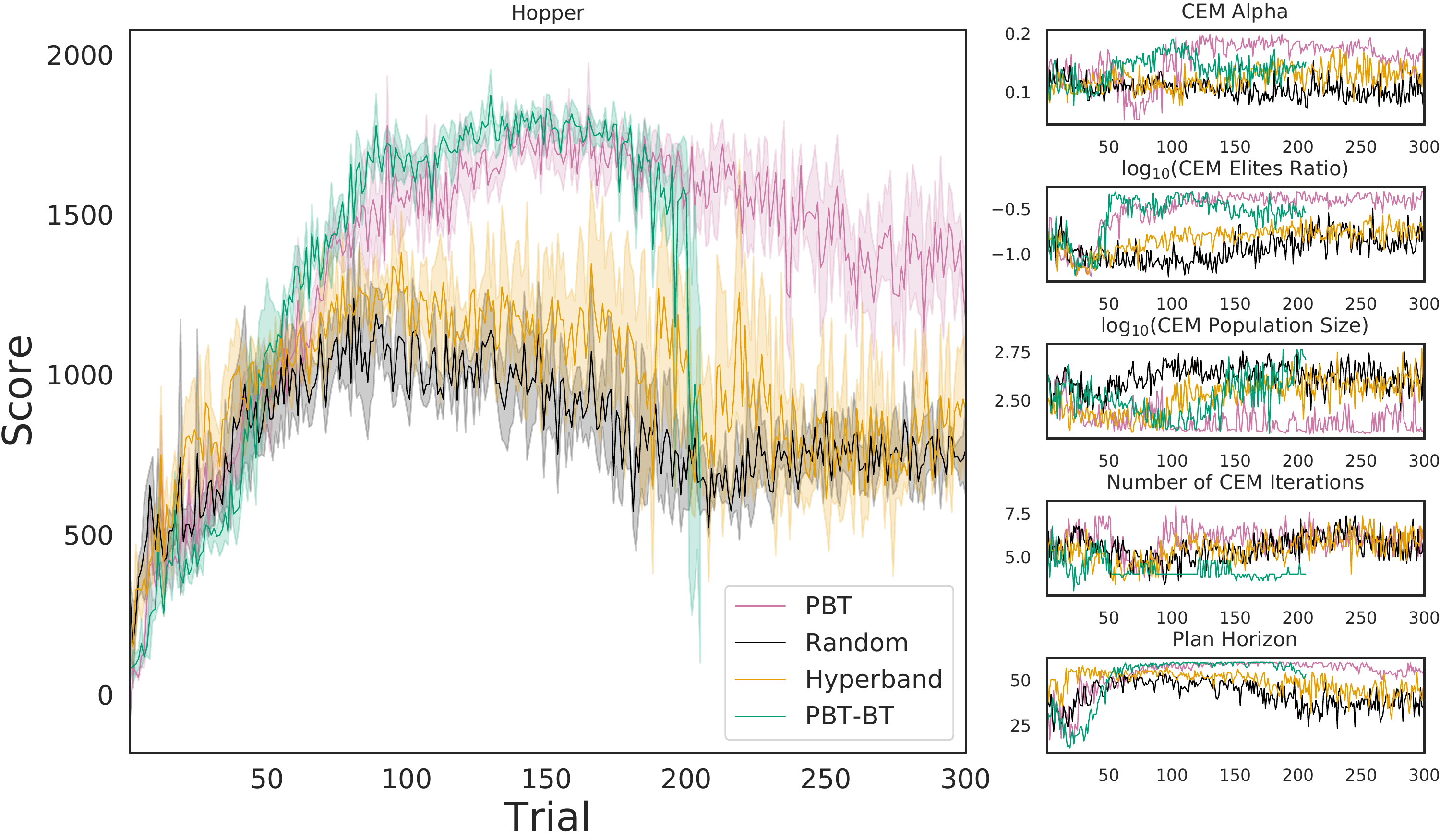}\label{fig:pbt_vs_rs_res_hopper_cem}}
    \subfigure[HalfCheetah CEM Optimizer]{\includegraphics[width=0.43\textwidth]{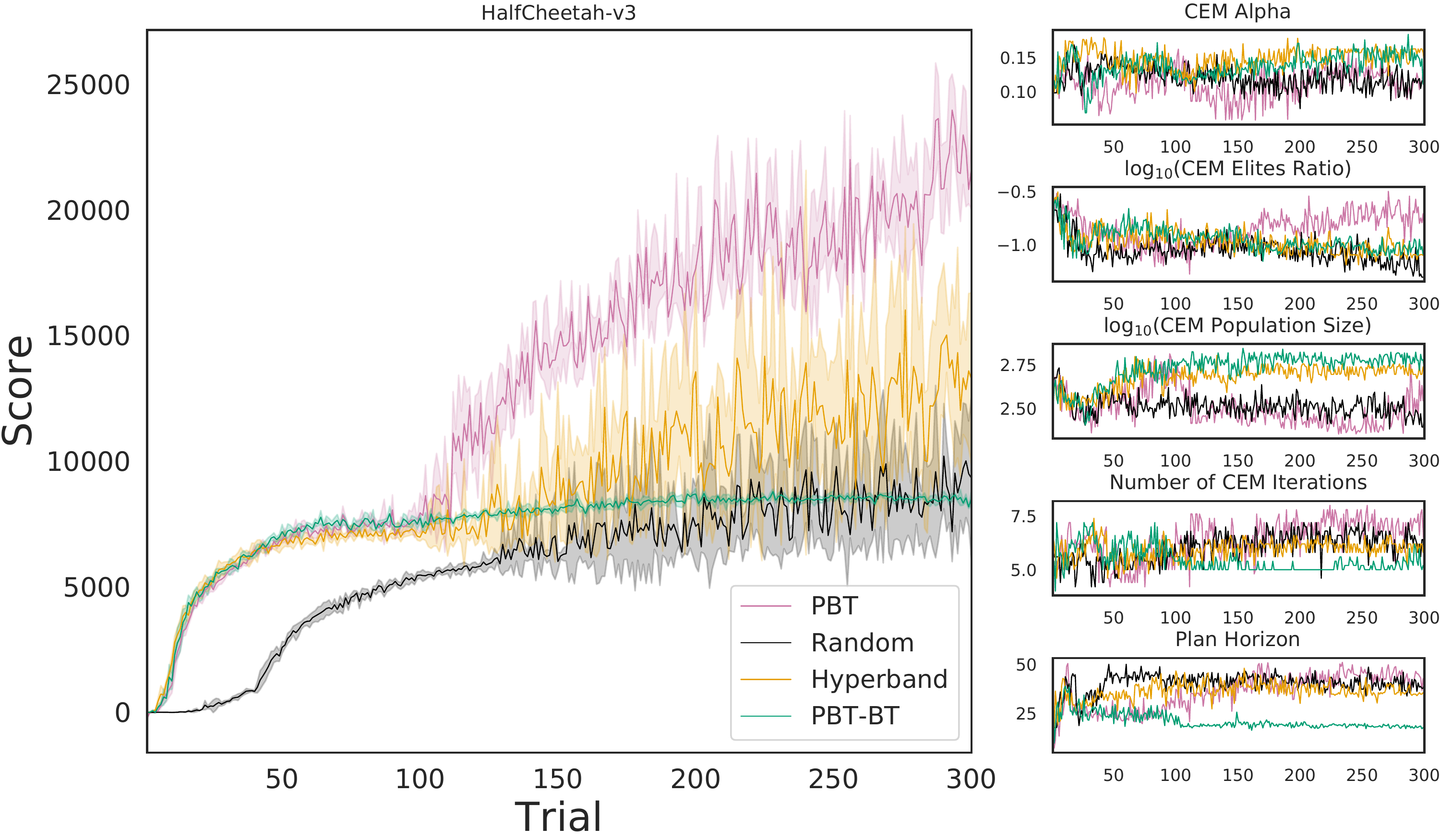}\label{fig:pbt_vs_rs_res_chee_cem}}
    \hfill
    \subfigure[Hopper Model Training]{\includegraphics[width=0.43\textwidth]{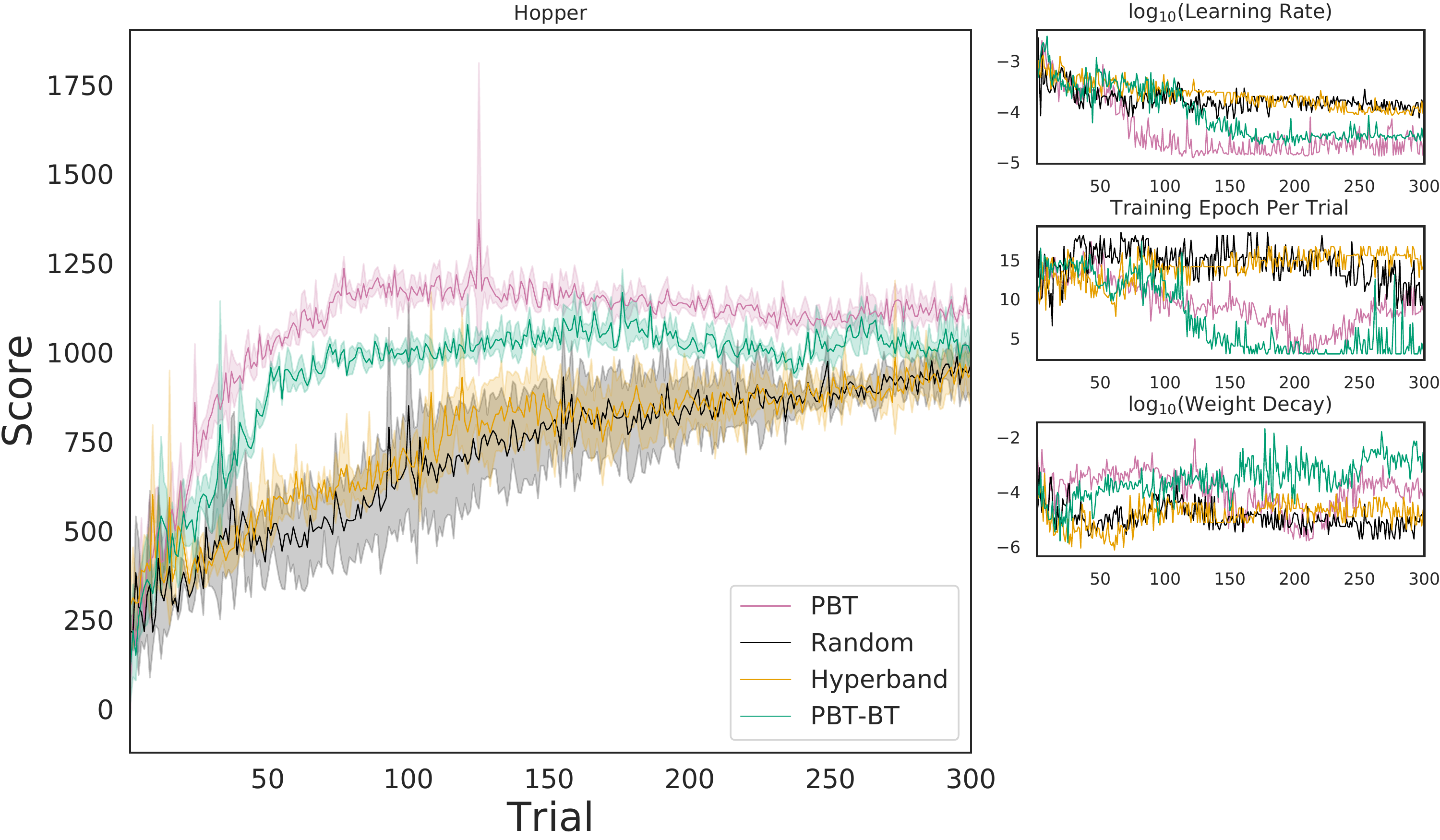}\label{fig:pbt_vs_rs_res_hopper_model}}
    \subfigure[HalfCheetah Model Training]{\includegraphics[width=0.43\textwidth]{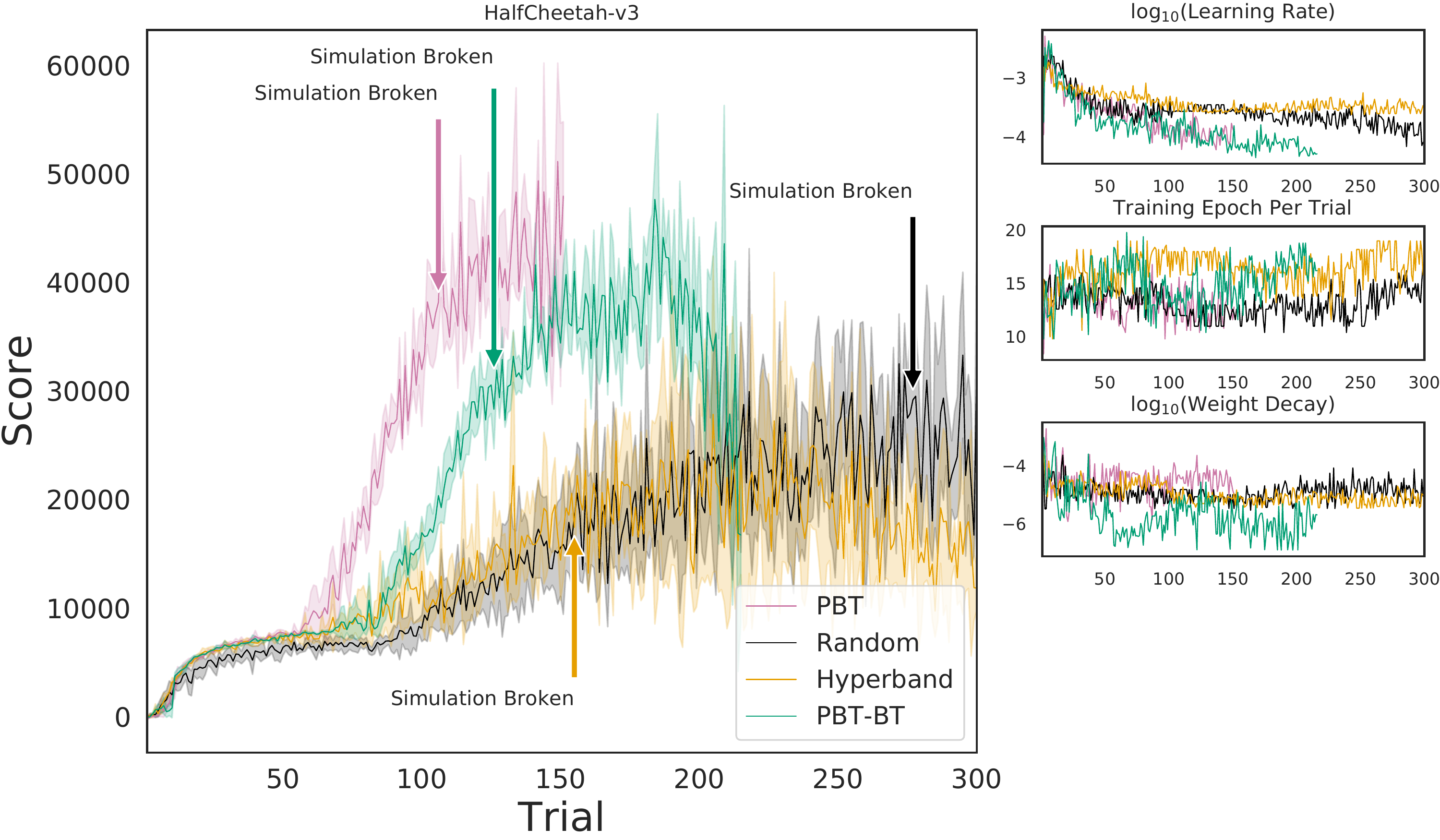}\label{fig:pbt_vs_rs_res_chee_model}}
\caption{Training curves for PBT, Hyperband and Random Search on Hopper and Halfcheetah. We reported the mean and standard deviation of the \emph{top $5$} members at each time step. On the $x$-axis we show the number of trials and on the $y$-axis the received reward. The stacked columns right of the learning curves show the average hyperparameters value of the top 5 members in the population at each time step, which indicates the trends of the hyperparameters across the training. 
Note, our algorithm discovers a bug in HalfCheetah environment which allows the agent's forward speed up to infinity and breaks the simulation. 
Therefore, we stop plotting the curves when $1/3$ members in the population stop training due to this error. PBT outperforms Random Search and Hyperband in all four environments during the search.
}
\label{fig:pbt_vs_rs_res}
\end{figure*}

When tuning the hyperparameters of PETS we observed two surprising results in the HalfCheetah environment.
The task for the agent in this environment is to control a ``Cheetah'' such that it runs as fast as possible to the right.
Before learning to run, the agent first has to learn to move right, then learn to walk and finally it can learn to run.
Besides those expected gaits some of our tuned agents were able to learn a new \emph{rolling} gait\footnote{Video provided in the supplementary material.}.
With this gait the Cheetah continually cartwheels and can build up far higher speeds than with running, allowing it to reach extremely high rewards. Some runs are so fast in fact that the simulator breaks with an overflow error (see Figures~\ref{fig:eval_compare} \& \ref{fig:pbt_vs_rs_res}).

\subsection{Effects of Hyperparameters}

We observed that the effect of HPO of model training vs CEM optimizer hyperparameters varied between the environments. 
While HPO of CEM hyperparameters led to more significant gains on Hopper, for HalfCheetah model parameters were more important to tune.

\paragraph{CEM Hyperparameters}
The impact on final performance of the CEM optimizer hyperparameters is not as large as that of model hyperparameters (see Figures~\ref{fig:hist_all_searches}). Comparatively fewer runs for both Hopper and HalfCheetah result in low rewards (around $0$) for the HPO of CEM hyperparameters, which shows that an inaccurate model will lead to catastrophic performance, whereas a weak planner will still be able to have mediocre performance. Tuning \textit{CEM optimizer} hyperparameters using PBT-BT, however does not allow agents to learn the rolling behaviour in HalfCheetah: most PBT-BT and random search agents result in a final score around $10\,000$.
This can be attributed to the backtracking mechanism, which replaces stumbling or falling agents, resulting in a worse reward, with ones that have a better running behavior.
Unexpectedly, falling and stumbling can be converted into the rolling gait, resulting in much higher rewards.

\paragraph{Model Training Hyperparameters} 
Figures~\ref{fig:hist_all_searches}c \& \ref{fig:hist_all_searches}d show that results vary widely between different hyperparameter configurations, and the score of the best configurations can be orders of magnitude larger than the score of the worst ones. 
Relying on static tuning methods, including random search or Hyperband, to tune the model parameters results in most evaluated configurations not being able to learn any significant behaviour, most often resulting in close to zero reward.
This suggests that the model parameters need to be chosen much more carefully to train a successful agent.

\subsection{Trends of Hyperparameters}
Figure~\ref{fig:pbt_vs_rs_res} depicts the learning performance of agents on the Hopper and HalfCheetah tasks.
Additionally, for each hyperparameter, we plot plausible hyperparameter schedules by taking the mean of the top 5 members' values for those hyperparameters.

\paragraph{Trends on Hopper}
We observe that dynamic tuning methods perform much better than static tuning methods for comparable budgets.
Regarding individual hyperparameters, we observe that the \textit{Plan Horizon} does not have a clear trend in the beginning but increases later for all methods. This might be due to the fact that when the model has been trained for more epochs, it becomes more reliable, allowing for better long-term planning. We discuss more on this in Appendix~\ref{sec:objective_mismatch}.
Similar to \citet{wang-arxiv19a} we observe that the best static configurations choose a final plan horizon between $20$ and $40$.
\textit{CEM Alpha} and \textit{CEM elites ratio} seem to have a clear mismatch between the schedules extracted from static and dynamic tuning methods. PBT and PBT-BT tend to be more greedy as these two values increase over time, whereas static tuning methods do not show this behavior. We also note that both PBT variants learn a decreasing learning rate schedule, similarly as reported by \citet{jaderberg2017population} without using any learning rate scheduler.

\begin{figure*}[t]
\centering
\includegraphics[width=0.94\linewidth]{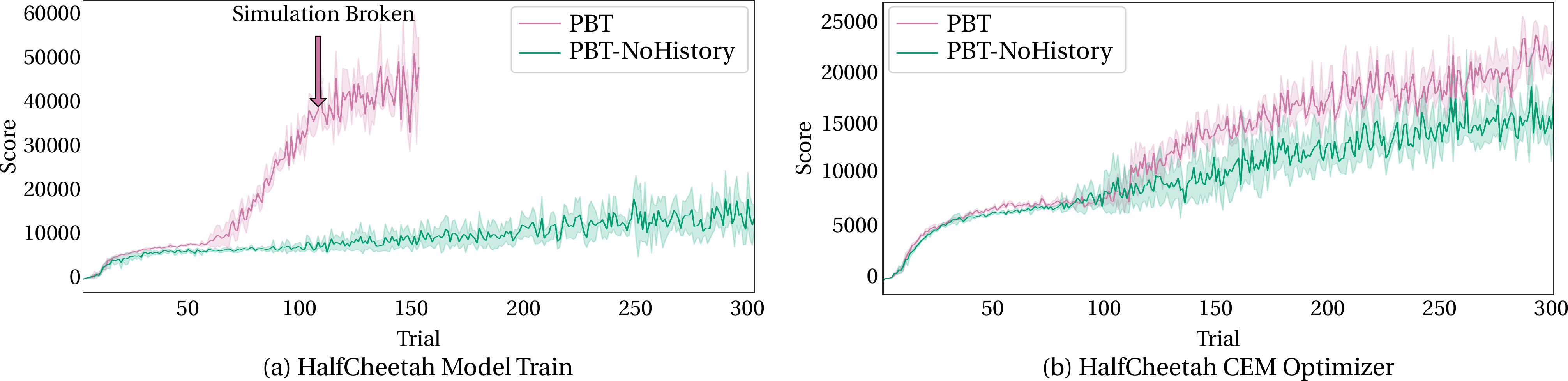}
\caption{PBT on HalfCheetah showing the average performance of the top $5$ members. The green curves are obtained by not copying the history when members are replaced by the top members. The performance drops 25\% and 75\% in HalfCheetah for \textit{CEM Optimizer} and \textit{Model Training} hyperparameters respectively, indicating the need for copying the history when applying PBT to MBRL algorithms}
\label{fig:history}
\end{figure*}

\paragraph{Trends on HalfCheetah}
On the HalfCheetah task as well, the optimization methods do not always find the same parameter schedules for the CEM hyperparameters. 
For example, all methods result in a mostly static value for \textit{CEM Alpha}, but converge to different values, whereas only PBT-BT over time slowly decreases the \textit{Plan Horizon}. Interestingly, PBT tends to reduce \textit{CEM Population Size}, which does not match our intuition. We can observe this phenomenon also in the Hopper environment.
Furthermore, we can observe that PBT is able to train agents which learn much faster. The large difference in score is because a few well performing agents are able to successfully learn the rolling gait while PBT-BT quickly prevents the ``Cheetah'' from falling over, a necessary precondition before learning to roll.
However, the whole PBT-BT population reliably learns a fast running gait. An interesting observation when jointly optimizing all the hyperparameters is that Random Search is better than PBT as can be seen in Appendix~\ref{appendix:joint}.
However, we can see that PBT successfully reaches an impressive episode return with around $50\;000$ for the top $5$ members, see Figure~\ref{fig:pbt_vs_rs_res_chee_model}. 
Two possible reasons are (i) during optimizing with PBT, good trajectories are propagated by PBT's exploitation mechanism which copies not only the weights of the neural network but also the hyperparameters of the network and (ii) PBT searches more in the region that gives better returns through its explore mechanism.
Again both PBT variants automatically adapt the learning rate.

\paragraph{Transferability of Hyperarameter Schedules}
The dynamic schedules learned on the environments are clearly not transferable across even runs while the static ones don't transfer across environments, as is shown in Figure~\ref{fig:eval_compare}. This is in stark contrast to dynamic runs usually giving the best performing results in those experiments.
As discussed in Section~\ref{sec:appendix}, environment specific configurations can allow an MBRL agent to learn a well performing dynamics model but this might not transfer well across runs or even at all to other environments.
This might be attributed to the intialization of RL agents, which differs from the intialization during the HPO run.
However, most search methods still outperform the hand tuned baselines for both sets of hyperparameters.
Only random search in Hopper with CEM Optimizer does not show a significant gain. One explanation may be that random search does not sample good configurations with the given budget.
Moreover, dynamic tuning schedules performed best in Hopper whereas in HalfCheetah, random search and Hyperband outperformed PBT and PBT-BT.
\subsection{Copying History across PBT Members}
As mentioned before, the original PBT paper~\citep{jaderberg2017population} does not address the importance of the history/replay buffer, which is critical in MBRL. We performed further experiments to shed light on this aspect and discovered that it is crucial to the performance (see Figure~\ref{fig:history}). When not copying the history to other members during the exploitation step of PBT, the performance has significant drops for both sets of hyperparameters. This also reveals that in the context of MBRL, the history has a significant impact on model training and implicitly influences the planning part.
\paragraph{Summary of Experimental Results}
Based on this empirical evaluation we revisit the choices (see Section~\ref{sec:approach}) MBRL practicioners have to face before making use of HPO.
To this end, for the PETS method we suggest to make use of dynamic tuning, if the most important evaluation criterion is final performance.
If however, a robust configuration that is transferable across multiple runs is required, static configuration provides the desired result. Moreover, the history is an essential factor for MBRL during training. When applying PBT to MBRL, we suggest to also copy the history during the exploitation step.

%% file: 6_conclusion.tex
In this paper, we thoroughly investigated the applicability of HPO to MBRL and explored different design choices MBRL practitioners have to make when tuning hyperparameters.
We empirically evaluated the influence of these design choices using HPO on the PETS algorithm, and observed not only that different hyperparameters in MBRL can drastically influence the final performance, but also that HPO significantly outperforms manually tuned hyperparameter settings.
Our experiments support the conclusion that automatic hyperparameter optimization can help researchers in finding settings that allow agents to learn faster and achieve better final rewards aross different tasks.
In addition, we evaluated the use of dynamic hyperparameters that change through the learning process and observed that they clearly outperform static hyperparameters configurations (e.g., for \textit{plann horizon}).

We believe that our results validate the significant impact of hyperparameter tuning on learning performance. 
As a result, we want to encourage the MBRL community to apply automatic configuration methods, and especially dynamic tuning methods to reduce the need for human-expertise and to take into account the non-stationarity of the RL problem.
Exciting future research includes the joint and dynamic search for hyperparameters together with neural architectures, and the automatic search of alternative surrogate losses with HPO to further improve data-efficiency of MBRL.

%% file: 7_acknowledgement.tex
Andr\'e, Baohe, Frank and Raghu gratefully acknowledge support by the Bosch Center for Artificial Intelligence, and by the European Research Council (ERC) under the European Union’s Horizon 2020 research and innovation program under grant no. 716721, as well as by the German Research Foundation (DFG) through grant no INST 39/963-1 FUGG. They would like to thank their group for helpful feedback and discussions. Raghu additionally gratefully acknowledges support by the BMBF grant DeToL. 
The authors would like to thank the reviewers for their time and effort for reviewing and improving the paper.

%% file: 8_appendix.tex
\subsection{Environment Overview}\label{appendix:env}
\begin{figure}[h]
    \centering
    \subfigure[Pusher]{\includegraphics[width=0.18\textwidth]{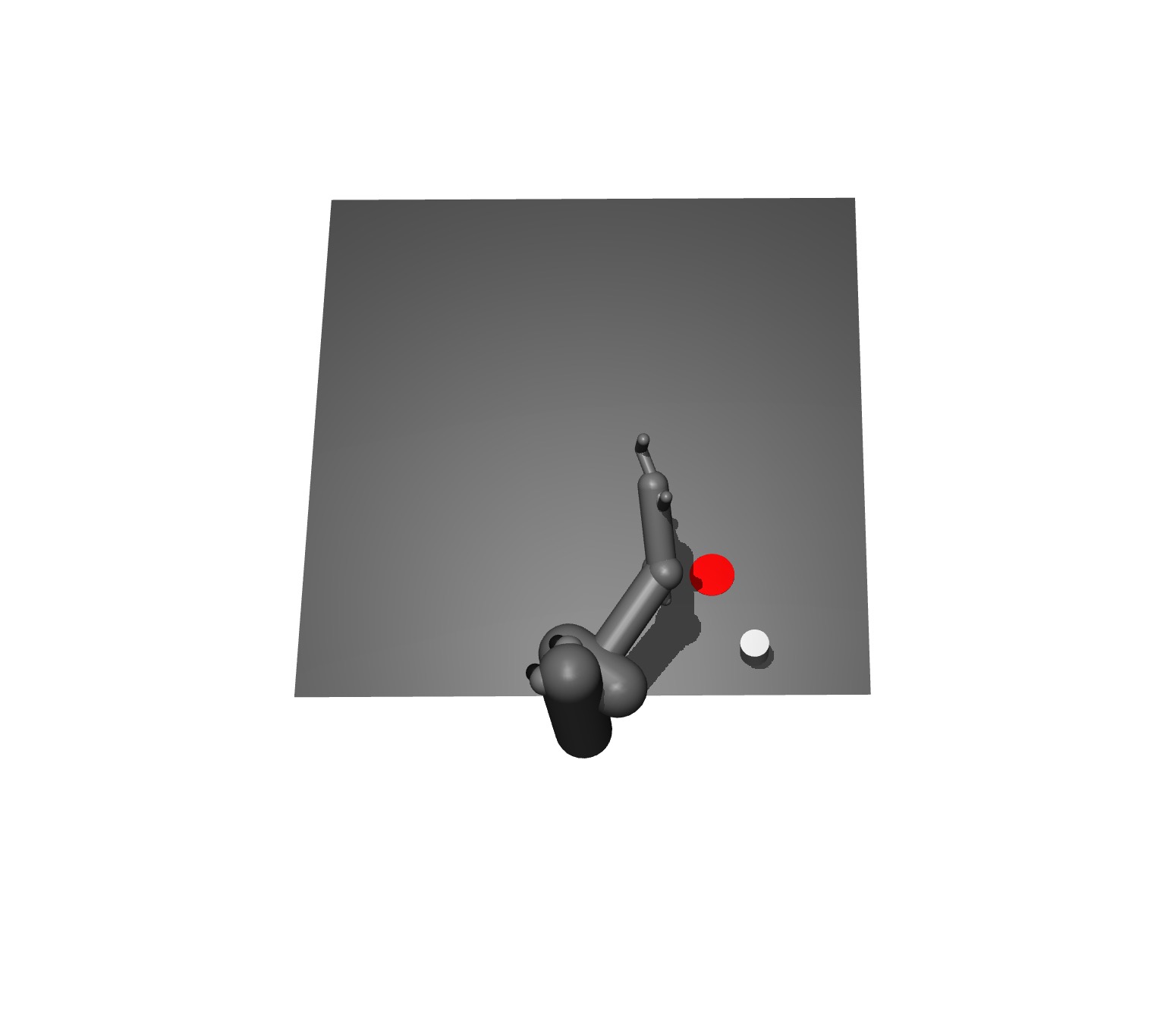}} 
    \subfigure[Reacher]{\includegraphics[width=0.18\textwidth]{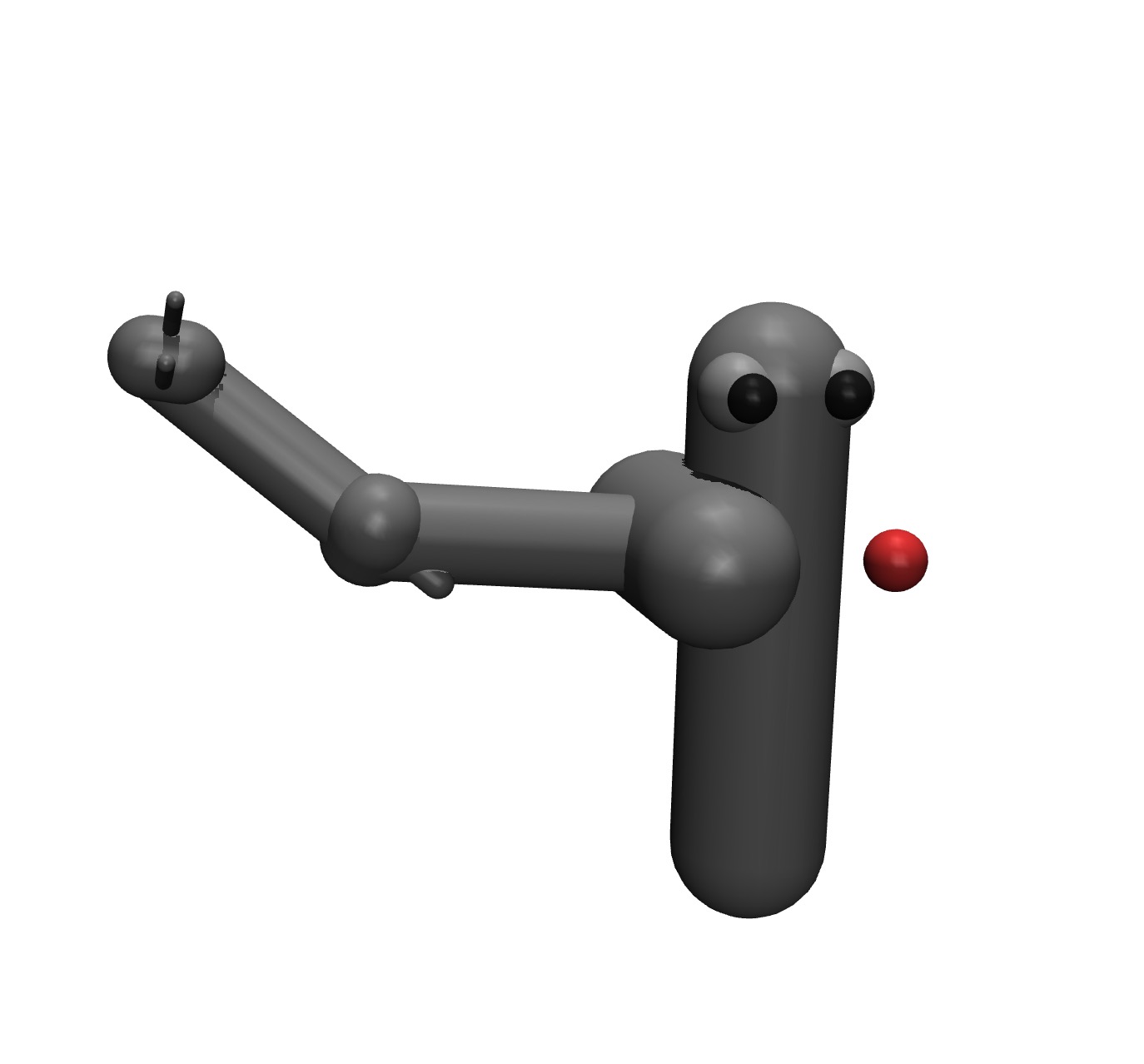}}
    \subfigure[Hopper]{\includegraphics[width=0.18\textwidth]{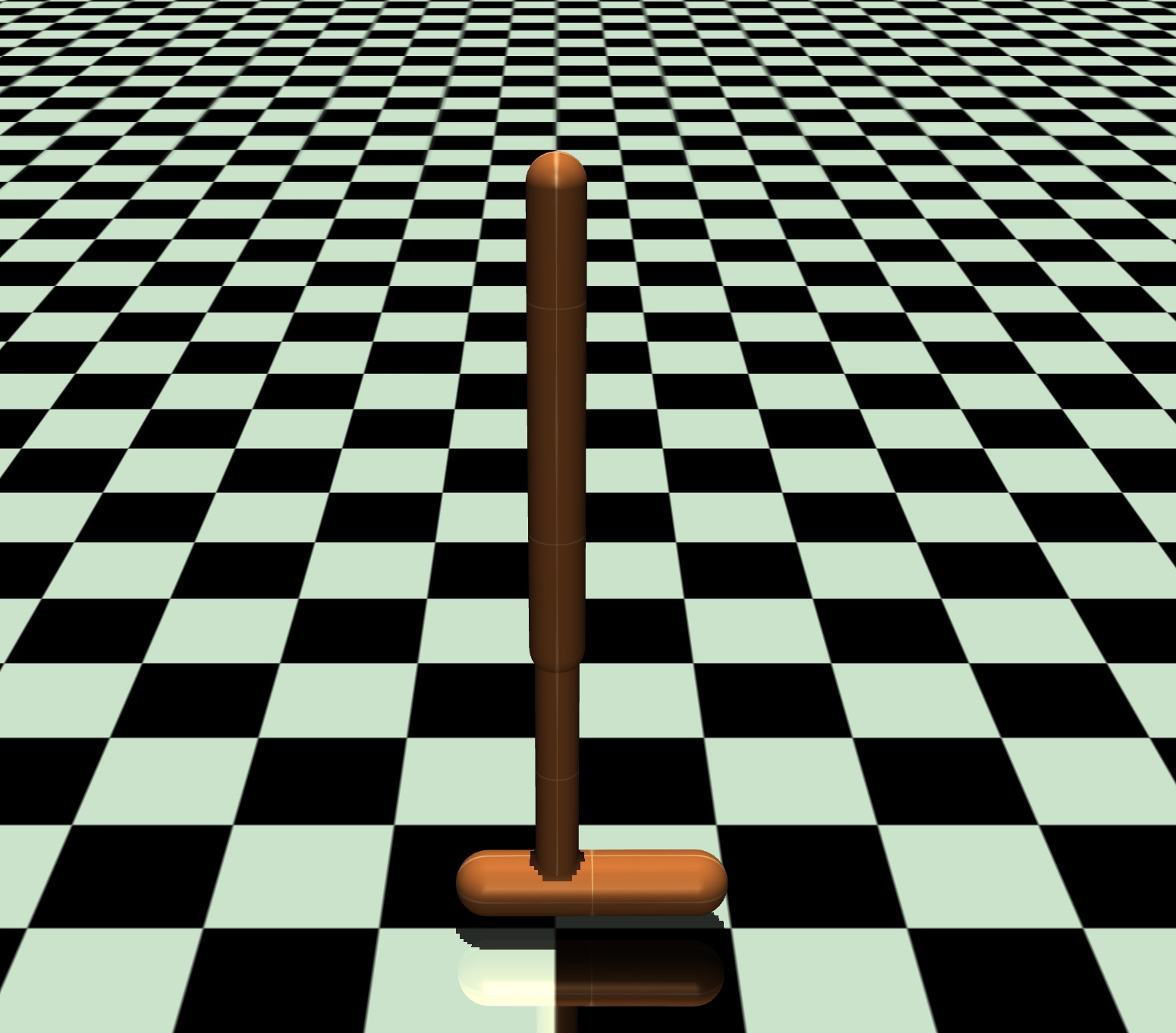}} 
    \subfigure[HalfCheetah]{\includegraphics[width=0.18\textwidth]{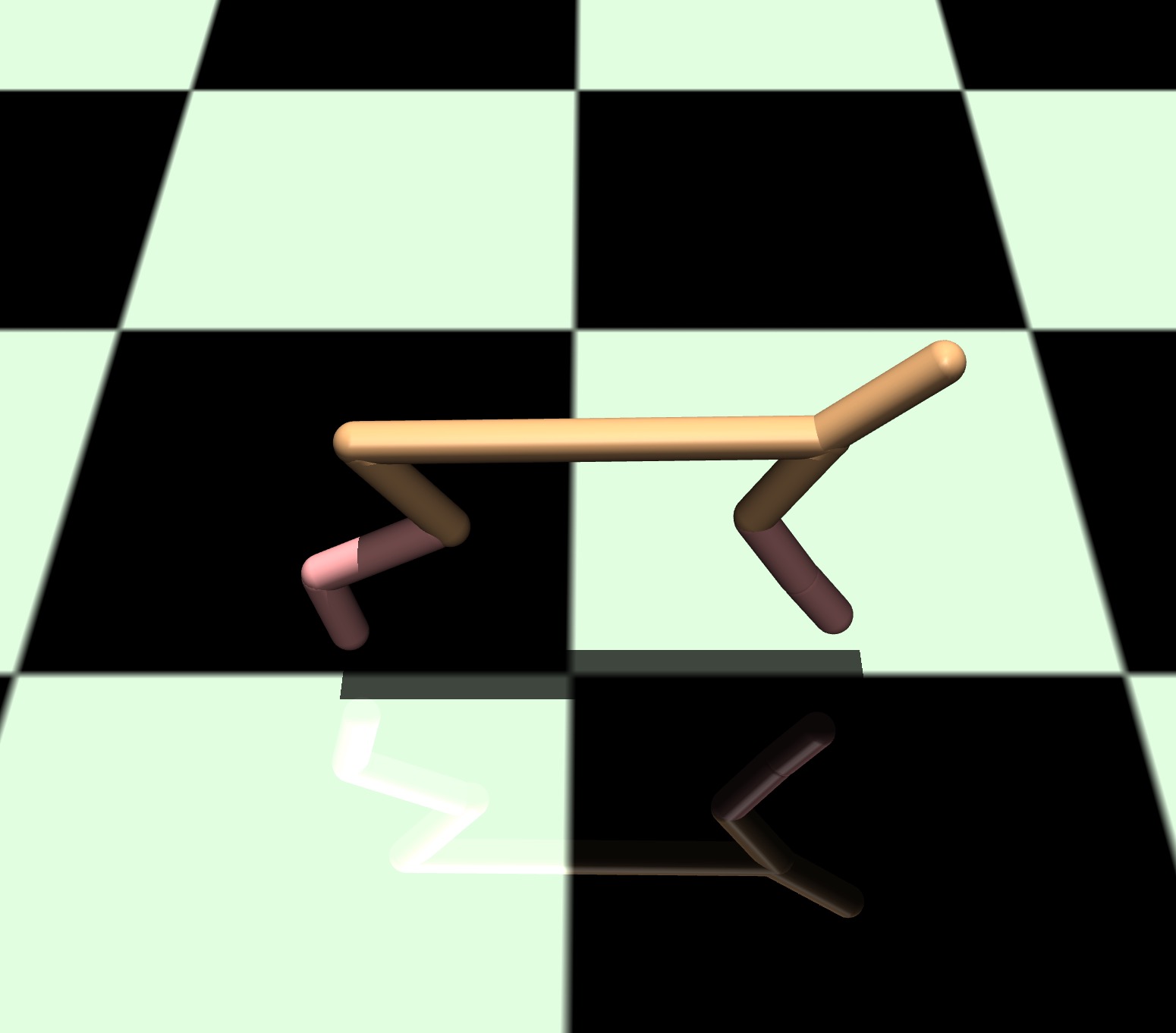}}
    \subfigure[Daisy]{\includegraphics[width=0.18\textwidth]{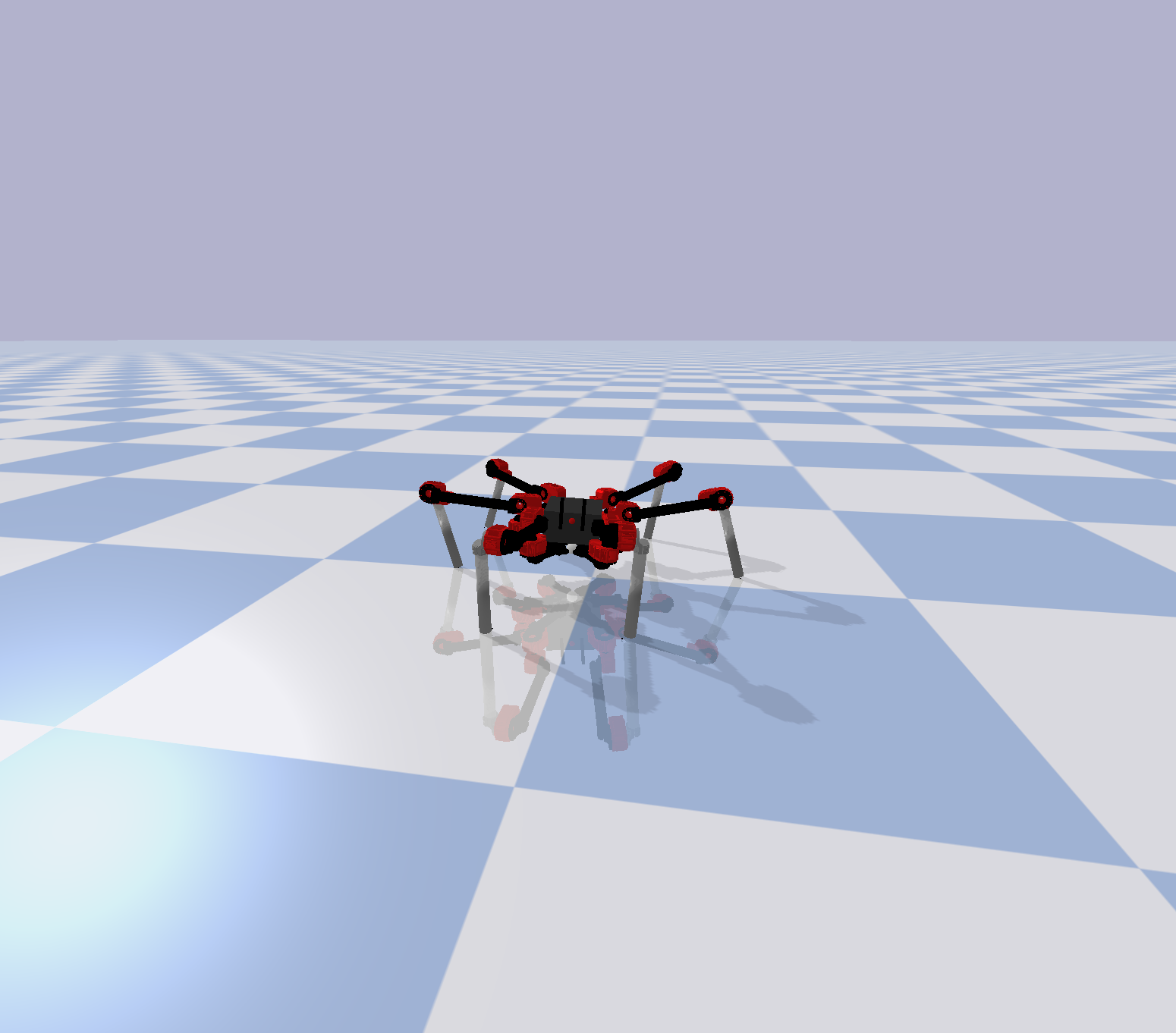}}
    \caption{Test environments used for the experiments}
    \label{fig:envs}
\end{figure}

We summarize the dimensionality, task horizon and the number of trials for experiments on each environment in Table~\ref{tab:env_setting}. Figure~\ref{fig:envs} shows a screenshot of each of the environments. 
\begin{table}[h]
\centering
\begin{tabular}{@{}c|c|c|c|c@{}}
\toprule
Name &
  \begin{tabular}[c]{@{}c@{}}Observation Space\\ Dimension\end{tabular} &
  \begin{tabular}[c]{@{}c@{}}Action Space \\ Dimension\end{tabular} &
  Horizon &
  Number of trials \\ \midrule
Pusher      & 20 & 7 & 150  & 80  \\
Reacher     & 17 & 7 & 150  & 80  \\
Hopper      & 12 & 3 & 1000 & 300 \\
HalfCheetah & 18 & 6 & 1000 & 300 \\
Daisy         & 24 & 18 & 1000 & 300 \\ \bottomrule
\end{tabular}
\caption[Dimensions of the observation and action space, task horizon and number of trials for all environments.]{\label{tab:env_setting} Dimensions of the observation and action space, task horizon and number of trials for all environments.
}
\end{table}

\begin{algorithm}[h]
\SetAlgoLined
  \KwIn{budgets $b_{min}$ and $b_{max}$, $\eta$}
  $s_{max} = \lfloor \log_{\eta}{\frac{b_{max}}{b_{min}}} \rfloor$\;
  \For{$ s  \in  \{ s_{max}, s_{max}-1,  \cdots ,0 \} $}{
   $n = \lceil \frac{s_{max}+1}{s+1} \cdot \eta^{s} \rceil$\;
   sample n configurations $C$\;
   $b = \eta^{s} \cdot b_{max}$\;
   \While{$b \leq b_{max}$}{
     $L = evaluate(C, b)$\;
     $k = \lfloor n_i / \eta \rfloor$\;
     $C = top_{k}(C, L)$\;
     $b = \eta \cdot b$\;
    }
  }
\KwOut{Configuration with the smallest loss}
 \caption{Hyperband Algorithm}
\label{algo:hyperband}
\end{algorithm}

\subsection{Implementation Details of Search Strategies}\label{appendix:pbt-bt}
\paragraph{Random Search}
For the random search experiments, we sample 40 independent configurations for each experiment and run them in parallel.
Each configuration is evaluated using the full budget~(i.e. number of trials) and the total amount of trials are $3200$ and $12000$ for Pusher/Reacher and the other environments respectively.
\paragraph{Hyperband}
Hyperband randomly samples configurations and utilizes Successive Halving to quickly reject poorly performing configurations. Moreover, Hyperband sets the budget for each fidelity such that the amount of time cost in each stage is approximately the same under the assumption that the time cost will be linear proportional to the amount of budgets. Compared to random search, Hyperband is slower than random search by a constant factor at most because of its dynamical trade-off between exploration and exploitation. A detailed algorithm is shown in Algorithm~\ref{algo:hyperband}.
We use the implementation from \citet{DBLP:conf/icml/FalknerKH18}. The \textit{minimal/maximal} budgets for Pusher are $8/80$ and $33/300$. $\eta$ is equal to $3$. We set the \textit{number of iterations} to 15 in order to have a fair comparison with the other methods. This resulted in $3130$ and $11982$ trials for Pusher/Reacher and the other environments respectively.
\paragraph{PBT}
For PBT, we deploy exploitation and exploration every $4$ and $5$ trials for Pusher/Reacher and the other environments respectively. In the \textit{exploit} step, runs with performance in the bottom $20\%$ are replaced by runs from the top $20\%$. In the \textit{explore} step, for the replaced runs, $75\%$ perturb the hyperparameters and the remaining $25\%$ randomly resample from the search space.
\paragraph{PBT-BT}
We reuse the hyperparameters of PBT in the \textit{exploit} and \textit{explore} step. To perform backtracking for PBT, we maintain a population of elites across all the timesteps elapsed so far. These are then compared to the current population members after every $30$ PBT timesteps and poorly performing members are replaced with elite members sampled randomly from the elite population. When a member is replaced by an elite member, its hyperparameters are modified in such a way as to ensure no other members reuse the same hyperparameters. We also relax the restriction of standard PBT in which members are only exploited by the other members which are in the same time step. PBT-BT can also be regarded as a more greedy version of PBT, since it performs additional exploitation steps using backtracking to elite members.

\subsection{Search Space}\label{appendix:searchspaces}
We use the same search space for all environments to ensure a fair comparison. Tables~\ref{pusher_sp} \&\ref{reacher_sp} \& \ref{hf_sp} show the considered search spaces (for Pusher \& Reacher \& Hopper, HalfCheetah and Daisy respectively) where the default value follows \citep{chua2018deep} as our baseline.

\begin{table}[h]
\centering
\begin{tabular}{|l|c|c|c|c|}
\bottomrule
Catagory                       & Hyperparameter           & Range            & \multicolumn{1}{l|}{Default Value} & Log-Transform \\ \toprule\bottomrule
\multirow{3}{*}{Model Train}   & Learning Rate            & {[}$3$e-$5$, $3$e-$3${]} & $1$e-$3$                               & True          \\ \cline{2-5} 
                               & Weight Decay             & {[}$1$e-$7$, $1$e-$1${]} & $4$e-$4$                               & True          \\ \cline{2-5} 
                               & Training Epochs          & {[}$3$, $20${]}      & $5$                                  & False         \\ \toprule\bottomrule
\multirow{5}{*}{CEM Optimizer} & Number of CEM Iterations & {[}$3$, $8${]}       & $5$                                  & False         \\ \cline{2-5} 
                               & CEM Population Size      & {[}$100$, $700${]}   & $500$                                & True          \\ \cline{2-5} 
                               & CEM Alpha                & {[}$0.01$, $0.5${]}  & $0.1$                                & False         \\ \cline{2-5} 
                               & CEM Elites Ratio         & {[}$0.04$, $0.5${]}  & $0.1$                                & True          \\ \cline{2-5} 
                               & Plan Horizon             & {[}$5$, $40${]}      & $30$                                 & False         \\ \toprule
\end{tabular}
\caption{\label{pusher_sp} Search space of PBT, Hyperband and random search for Pusher}
\end{table}

\begin{table}[h]
\centering
\begin{tabular}{|l|c|c|c|c|}
\bottomrule
Catagory                       & Hyperparameter           & Range            & \multicolumn{1}{l|}{Default Value} & Log-Transform \\ \toprule\bottomrule
\multirow{3}{*}{Model Train}   & Learning Rate            & {[}$1$e-$5$, $4$e-$2${]} & $7.5$e-$4$                               & True          \\ \cline{2-5} 
                               & Weight Decay             & {[}$1$e-$7$, $1$e-$1${]} & $5$e-$4$                               & True          \\ \cline{2-5} 
                               & Training Epochs          & {[}$3$, $20${]}      & $5$                                  & False         \\ \toprule\bottomrule
\multirow{5}{*}{CEM Optimizer} & Number of CEM Iterations & {[}$4$, $6${]}       & $5$                                  & False         \\ \cline{2-5} 
                               & CEM Population Size      & {[}$200$, $700${]}   & $400$                                & True          \\ \cline{2-5} 
                               & CEM Alpha                & {[}$0.05$, $0.4${]}  & $0.1$                                & False         \\ \cline{2-5} 
                               & CEM Elites Ratio         & {[}$0.04$, $0.5${]}  & $0.1$                                & True          \\ \cline{2-5} 
                               & Plan Horizon             & {[}$5$, $40${]}      & $25$                                 & False         \\ \toprule
\end{tabular}
\caption{\label{reacher_sp} Search space of PBT, Hyperband and random search for Reacher}
\end{table}

\begin{table}[ht]
\centering

\begin{tabular}{|l|c|c|c|c|}
\bottomrule
Catagory                       & Hyperparameter           & Range            & \multicolumn{1}{l|}{Default Value} & Log-Transform \\ \toprule\bottomrule
\multirow{3}{*}{Model Train}   & Learning Rate            & {[}$1$e-$5$, $4$e-$2${]} & $1$e-$3$                               & True          \\ \cline{2-5} 
                               & Weight Decay             & {[}$1$e-$7$, $1$e-$1${]} & $7.5$e-$5$                             & True          \\ \cline{2-5} 
                               & Training Epochs          & {[}$3$, $20${]}      & $5$                                  & False         \\ \toprule\bottomrule
\multirow{5}{*}{CEM Optimizer} & Number of CEM Iterations & {[}$3$, $8${]}       & $5$                                  & False         \\ \cline{2-5} 
                               & CEM Population Size      & {[}$200$, $700${]}   & $500$                                & True          \\ \cline{2-5} 
                               & CEM Alpha                & {[}$0.05$, $0.2${]}  & $0.1$                                & False         \\ \cline{2-5} 
                               & CEM Elites Ratio         & {[}$0.04$, $0.5${]}  & $0.1$                                & True          \\ \cline{2-5} 
                               & Plan Horizon             & {[}$5$, $60${]}      & $30$                                 & False         \\ \toprule
\end{tabular}
\caption{Search space of PBT, Hyperband and random search for Hopper, HalfCheetah, Daisy}
\label{hf_sp}
\end{table}

\subsection{Joint Optimization of All Hyperparameters}
\label{appendix:joint}
\begin{figure*}[h]
\centering
    \subfigure[Hopper Joint]{\includegraphics[width=0.43\textwidth]{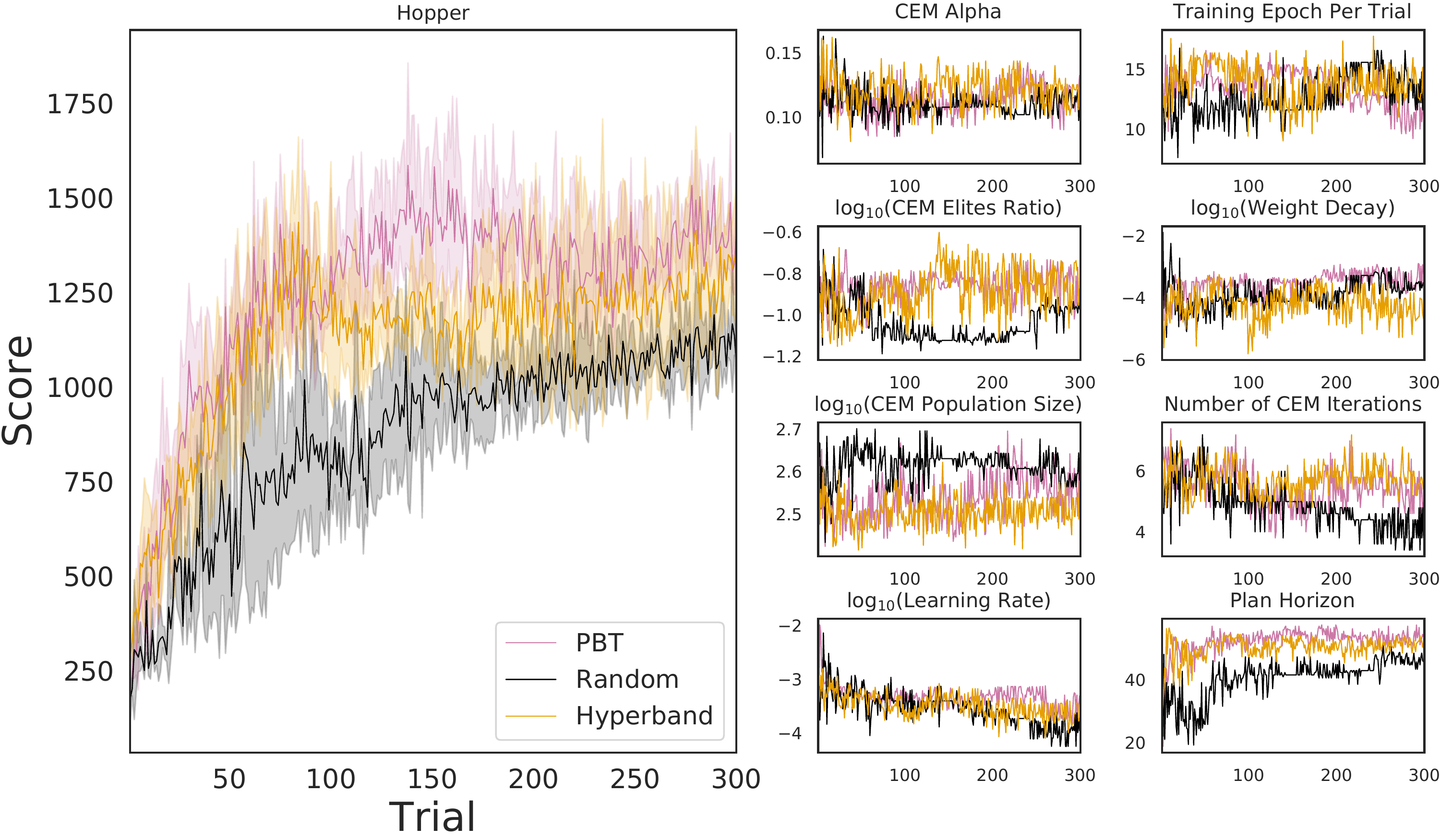}\label{fig:joint_hopper}}
    \subfigure[HalfCheetah Joint]{\includegraphics[width=0.43\textwidth]{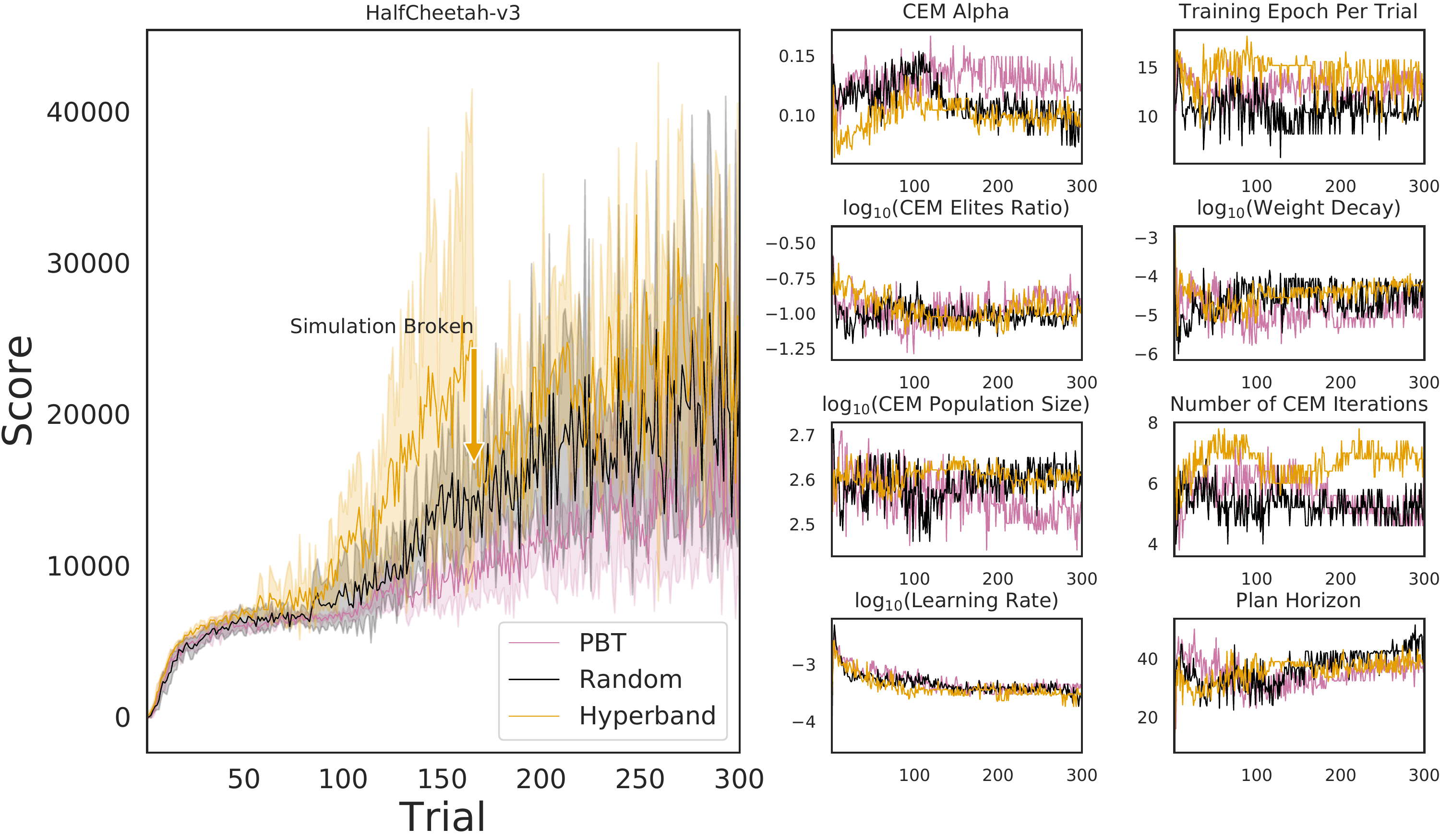}\label{fig:joint_halfcheetah}}

\caption{Joint optimization of both \textit{Model training} and \textit{CEM Optimizer} hyperparameters on Hopper and HalfCheetah using different search strategies. We show the average performance of the top $5$ members.}
\label{fig:joint}
\end{figure*}

In this section, we present the results obtained by jointly optimizing both sets of hyperparameters, namely \textit{Model Training} and \textit{CEM Optimizer}. Since the search space is larger than separately optimizing on two sets of hyperparameters. We increase the budget from 40 configurations to 80 configurations for PBT and random search. For Hyperband, we increase the number of iterations from 15 to 30. Overall, we now doubled the computational cost compared to optimizing only one set of hyperparameters at a time. The results are summarized in Figure~\ref{fig:joint}. From that, we find that PBT still gives the best performance in Hopper. However, it is worse than other two static tuning methods on HalfCheetah. This may have been caused by the fact that PBT is based on evolutionary strategies which generally do not perform well on high-dimensional problems. Therefore, without enough compute budget or a good heuristic, PBT may not be able to find a promising schedule that can outperform the hyperparameter configurations found by static tuning methods. The trends of hyperparameters found during joint optimization are similar to those found by optimizing them separately, e.g. for the learning rate , plan horizon and model weight decay.

\subsection{Additional Experimental Results on Other Environments}
\label{appendix:results}

\begin{figure*}[h]
\centering
\includegraphics[width=\textwidth]{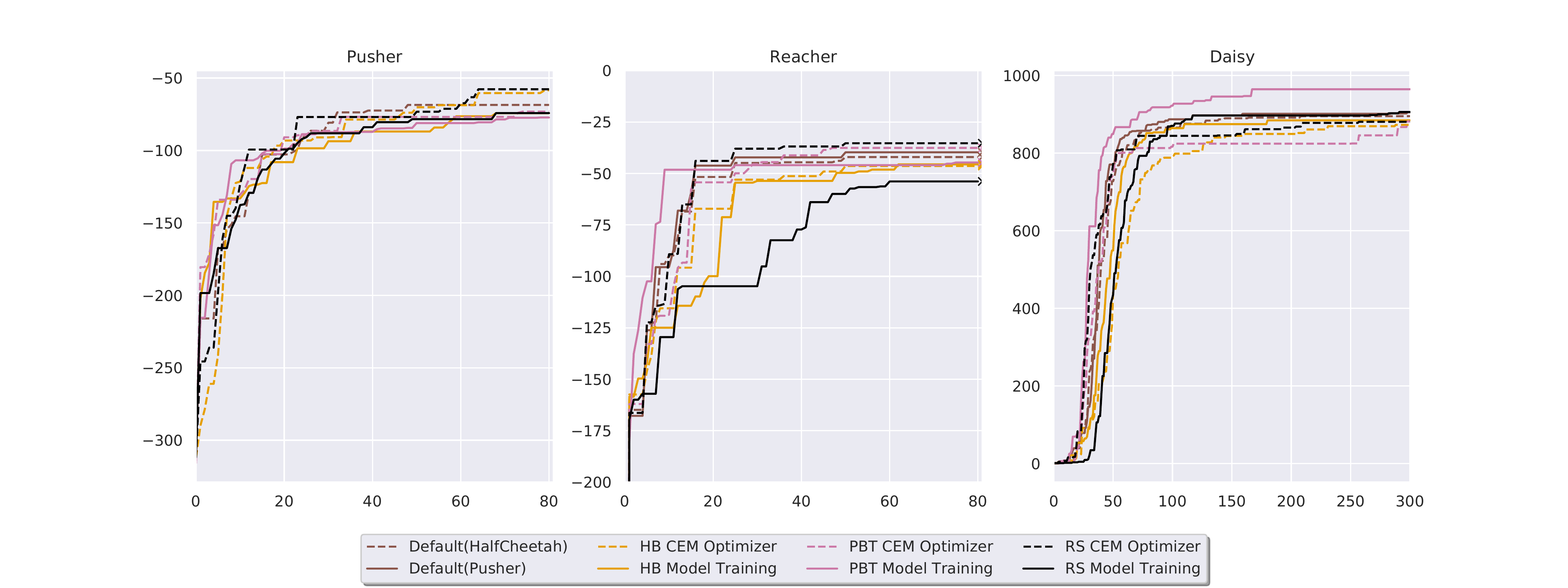}
\caption{Evaluation of the \emph{best} found hyperparameters and schedules from all search methods on Pusher, Reacher and Daisy. Scores~(y-axis) are the maximum of the average return over $5$ evaluations over number of trials~(x-axis). Brown lines are the baselines using the hyperparameters tuned by human expert.}
\label{fig:eval_compare_add}
\end{figure*}

\begin{figure*}[t]
\centering
    \subfigure[Pusher CEM Optimizer]{\includegraphics[width=0.43\textwidth]{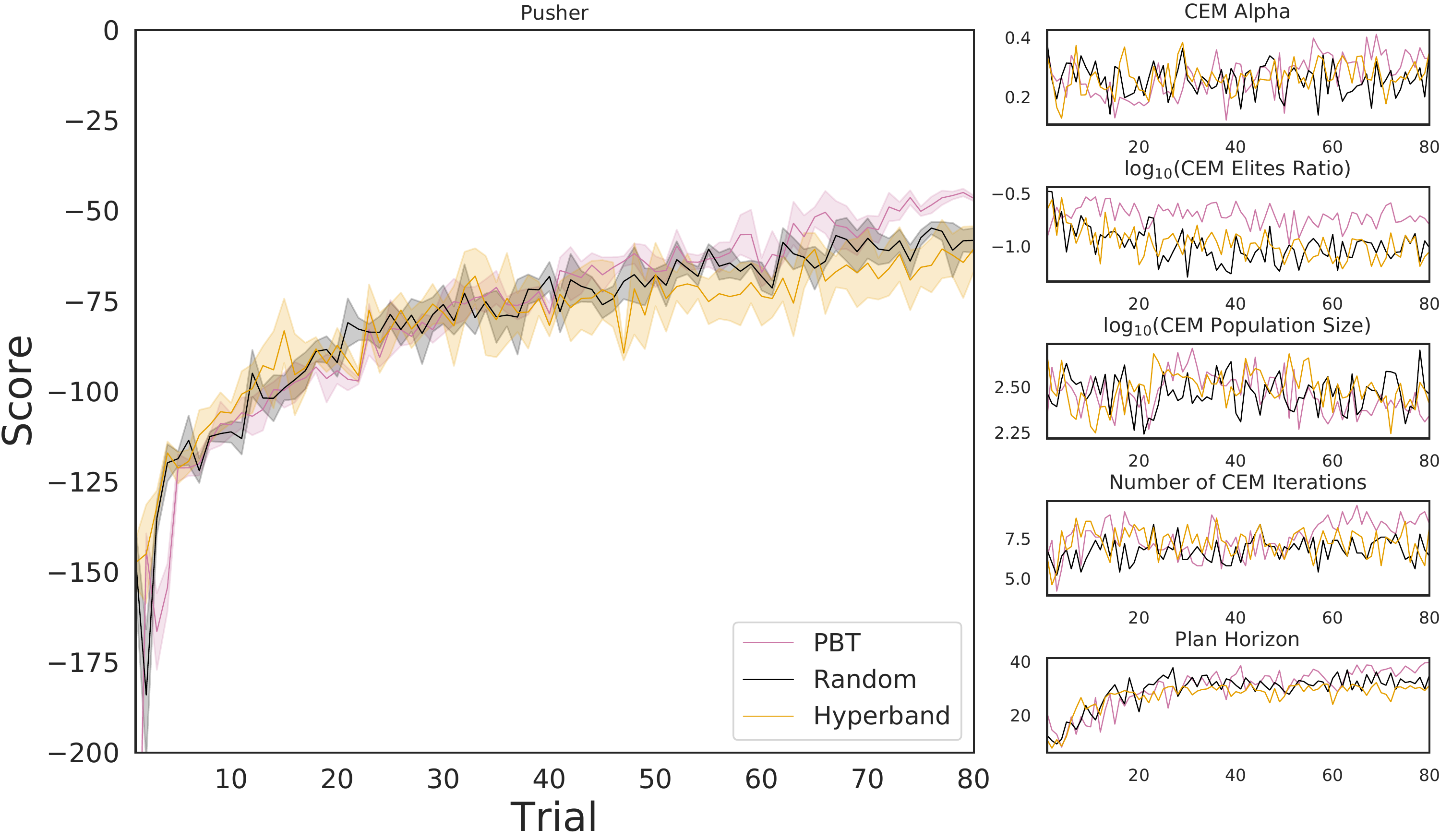}\label{fig:pbt_vs_rs_res_pusher_cem}}
    \subfigure[Pusher Model Training]{\includegraphics[width=0.43\textwidth]{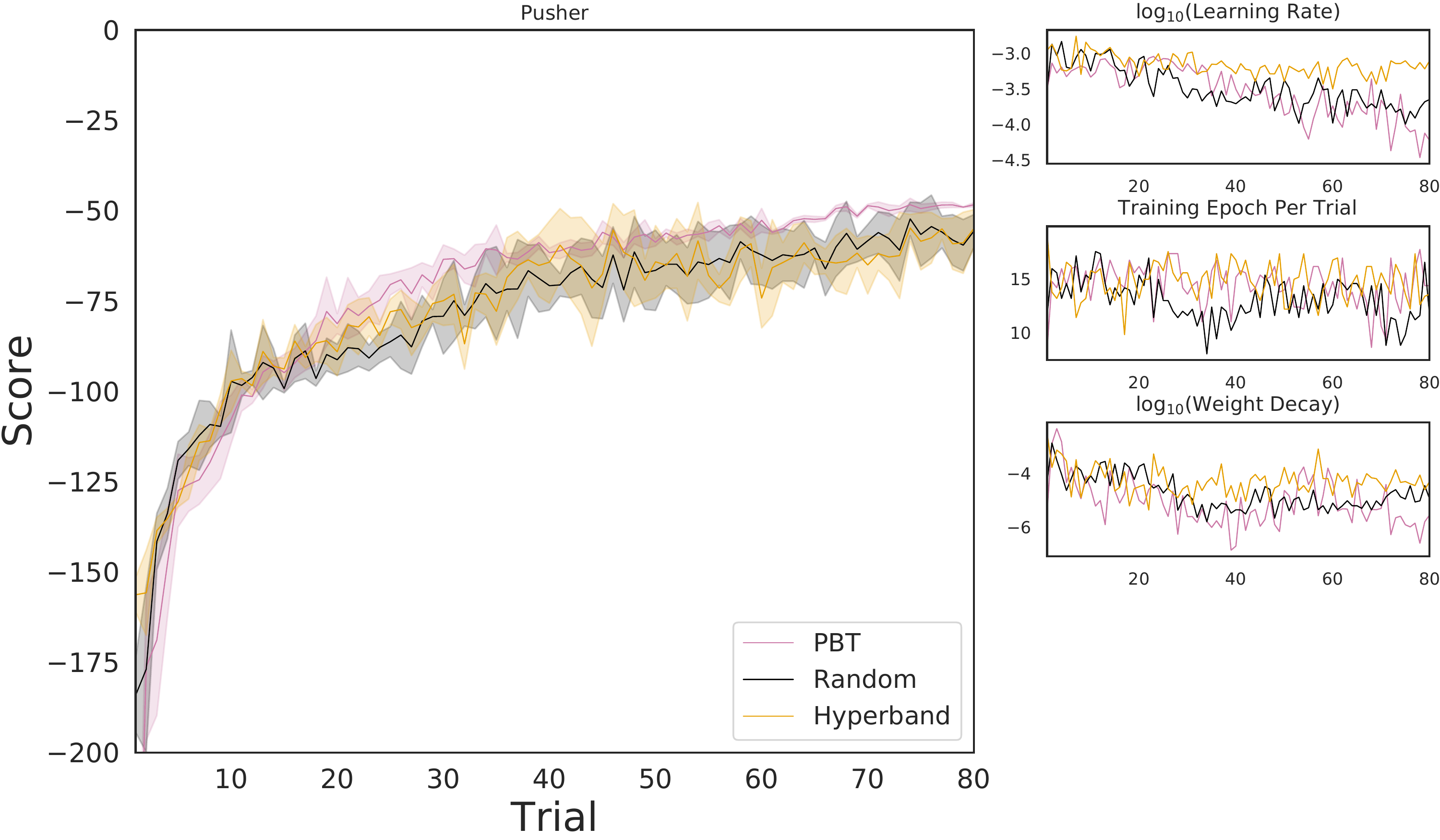}\label{fig:pbt_vs_rs_res_pusher_model}}
    \hfill
    
    \subfigure[Reacher CEM Optimizer]{\includegraphics[width=0.43\textwidth]{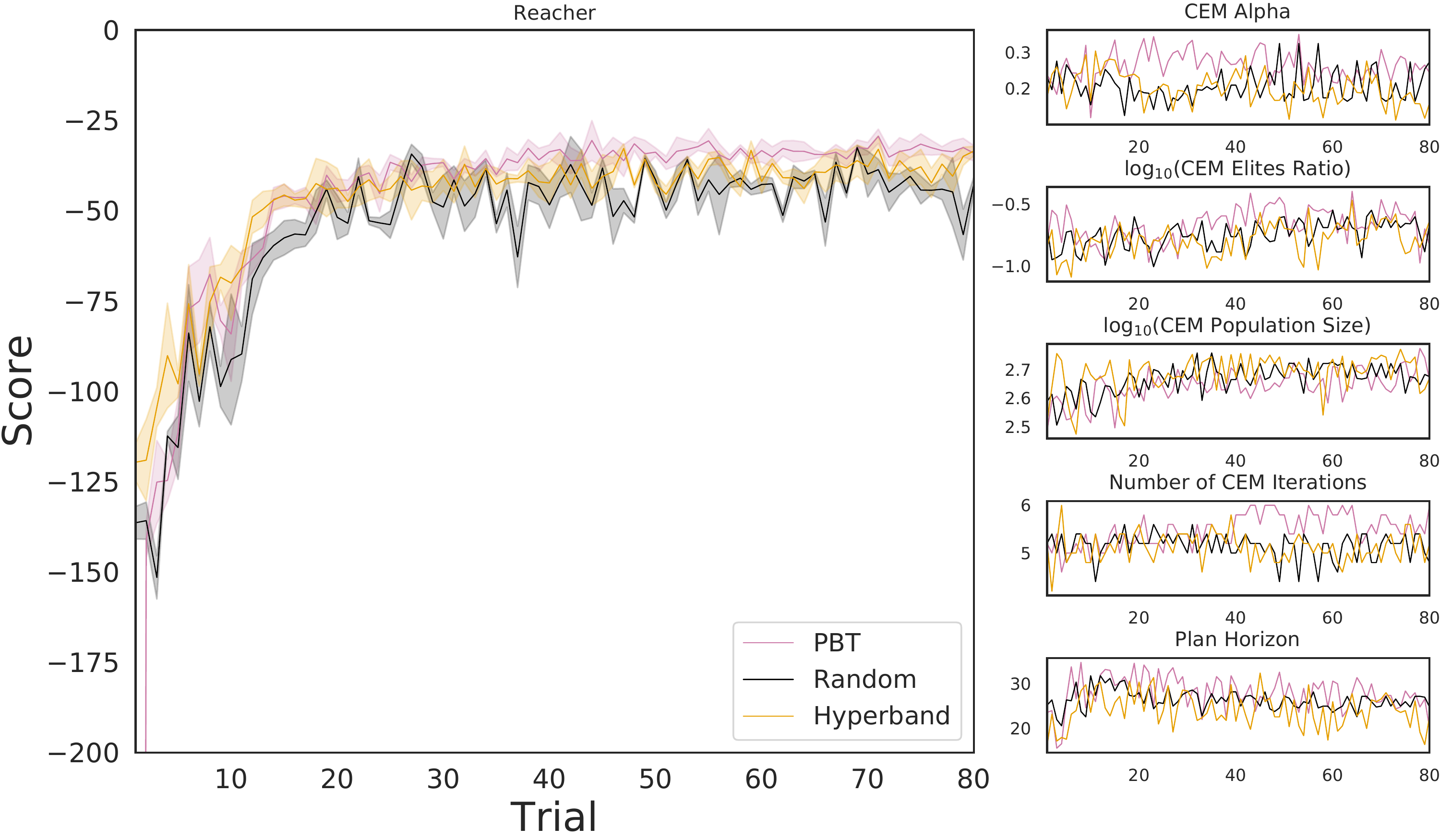}\label{fig:pbt_vs_rs_res_reacher_cem}}
    \subfigure[Reacher Model Training]{\includegraphics[width=0.43\textwidth]{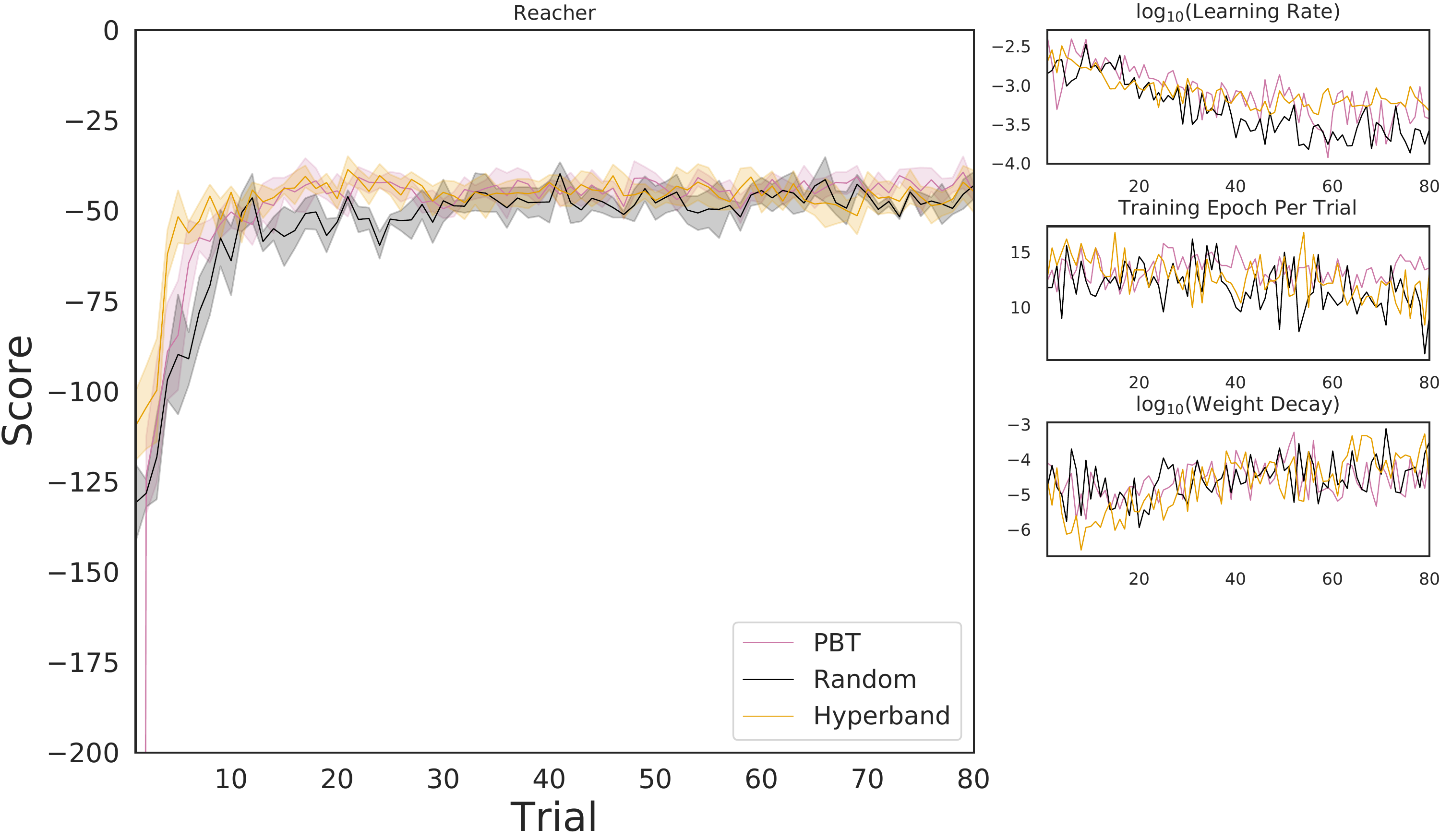}\label{fig:pbt_vs_rs_res_reacher_model}}
    \hfill
    \subfigure[Daisy CEM Optimizer]{\includegraphics[width=0.43\textwidth]{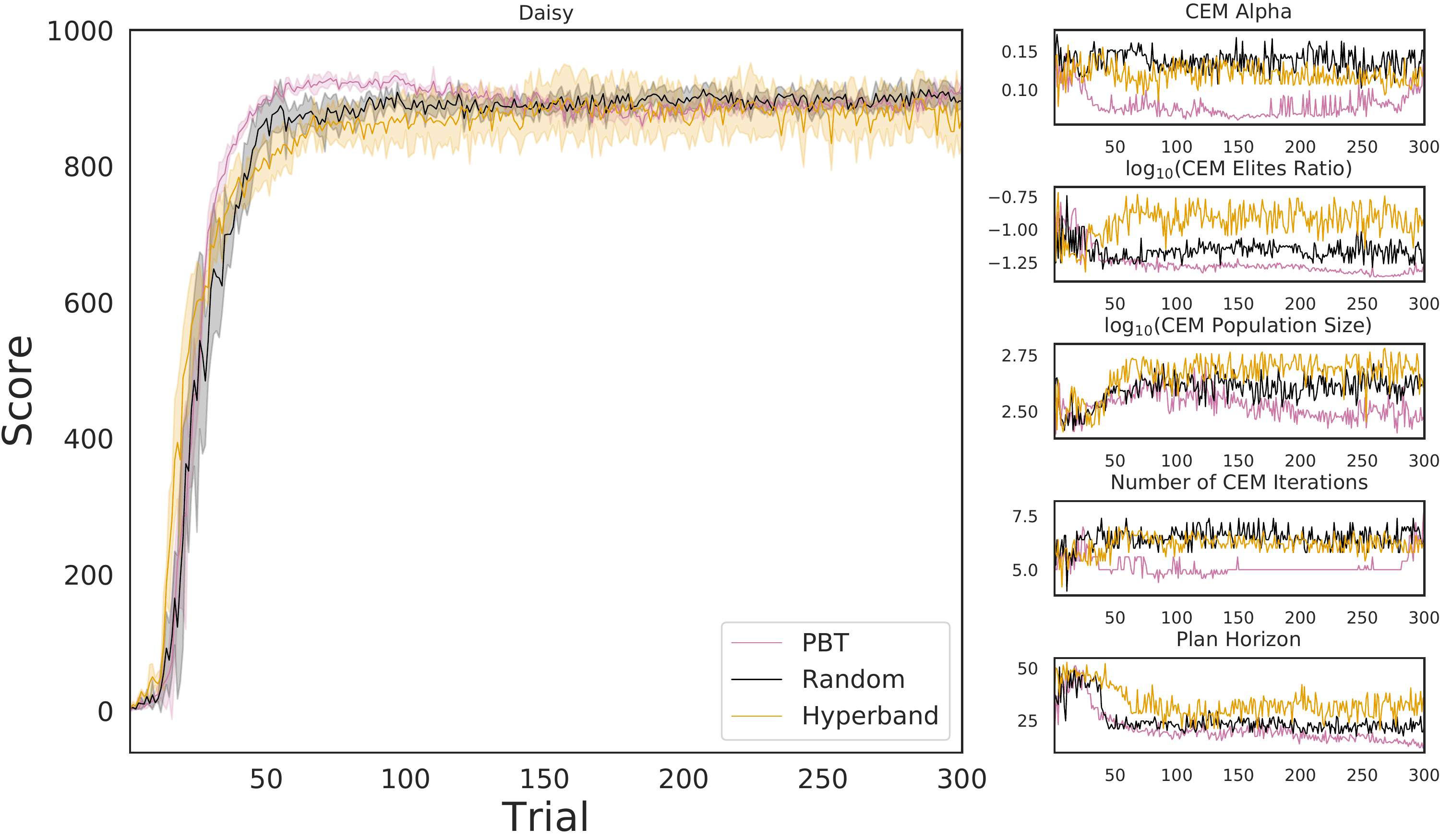}\label{fig:pbt_vs_rs_res_daisy_cem}}
    \subfigure[Daisy Model Training]{\includegraphics[width=0.43\textwidth]{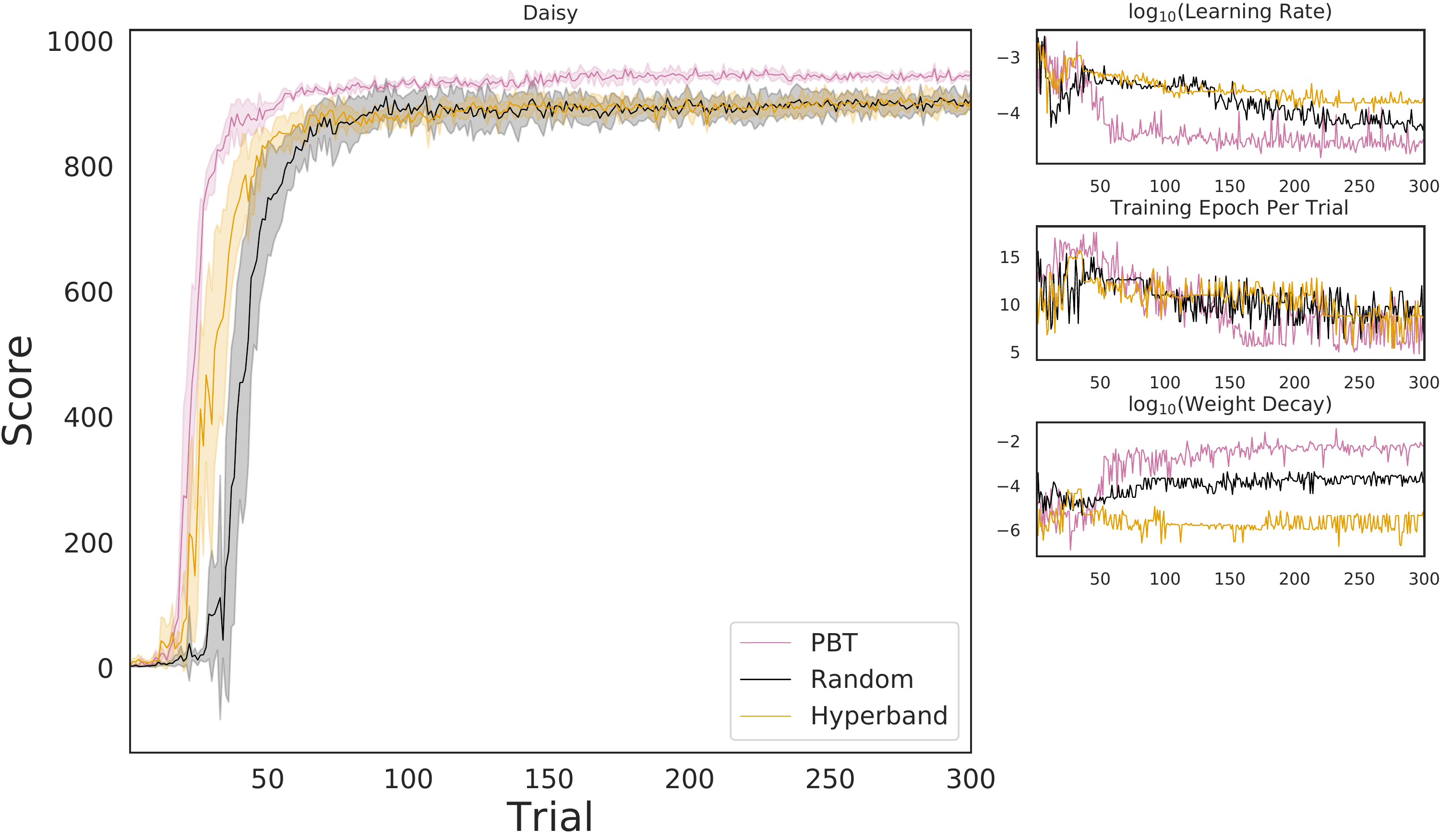}\label{fig:pbt_vs_rs_res_daisy_model}}
    
\caption{Training curves from PBT, Hyperband and Random Search on Pusher (top), and Reacher (middle) and Daisy (Bottom). We report the mean and standard deviation of the \emph{top $5$} members at each time step. On the $x$-axis we show the number of trials and on the $y$-axis the received reward. The stacked columns on the right of the learning curves show the average hyperparameter values of the top members in the population at each time step, which indicates the trend of the hyperparameters across training. 
}
\label{fig:pbt_vs_rs_res_add}
\end{figure*}

In addition to Hopper and HalfCheetah, we performed additional experiments on the Pusher, Reacher and Daisy environments. As is shown in Figure~\ref{fig:pbt_vs_rs_res} in the main paper, PBT-BT shows similar behavior in most cases and can get stuck in local optima. Therefore, we dropped PBT-BT to reduce the computational cost.
\paragraph{Pusher}
Since Pusher is a relatively simple task in our setting with a deterministic optimal final return given the initial state, there is only a small improvement in terms of the performance using PBT comparing to random search and Hyperband during the search, which is shown in Figures~\ref{fig:pbt_vs_rs_res_pusher_cem} \& \subref{fig:pbt_vs_rs_res_pusher_model}.
For all hyperparameters, the trends of them found by all methods are very close to each other.
Specifically, we found that all three strategies give the same increasing trend for \textit{Plan Horizon}, which again suggests that low plan horizon is more beneficial when the model is not yet fully trained.
The evaluation results of all hyperparameters and schedules are shown in Figure~\ref{fig:eval_compare_add}.
There we observe that the dynamic schedules learned by PBT have better performance in the initial training stage, which may be due to using a low plan horizon and larger learning rate in the initial stage.

\paragraph{Reacher}
Similar to Pusher, Reacher also has a deterministic final return and even lower dimensionality, which makes it easier to sovle compared to pusher. As a result, there isn't any improvement on applying dynamic tuning strategies since all three search strategies give similar results, which is shown in Figures~\ref{fig:pbt_vs_rs_res_reacher_cem} \& \subref{fig:pbt_vs_rs_res_reacher_model}.
Similar to Pusher, the trends for all hyperparameters found by all the methods are very similar.

\paragraph{Daisy}
Daisy is the hardest task in our setting due to its relatively high observation and action dimensionalities (see section~\ref{appendix:env}). Figures~\ref{fig:pbt_vs_rs_res_daisy_cem} \& \subref{fig:pbt_vs_rs_res_daisy_model} show the comparison of performances during the search and how hyperparameters evolve over time.
In this environment, in contrast to Pusher, PBT learns a strategy that \textit{decreases} \textit{Plan Horizon} over time.
This may suggest that for this task, long-term planning is not needed and could introduce cumulative noise in the model rollouts as it requires to perform one-step prediction with more iterations compared to short-term planning.
Therefore, a short-term greedy policy can be a good choice.
In terms of the \textit{Model Training} hyperparameters, PBT learns to increase the \textit{weight decay} over time, which indicates that it might be trying to combat overfitting which can otherwise occur for the model.
This is further supported by the decreasing trend of \textit{training epoch}.
We found that the evaluation of the hyperparameters and schedules does not cause as big an impact in the final return~(See Figure~\ref{fig:eval_compare_add}) as in the Hopper and HalfCheetah environments. However, the schedule of \textit{Model Training} found by PBT allows the agent to converge faster and results in better rewards than the other evaluated HPO methods. The baselines perform similar to the other search methods. This could be potentially caused by the design of the search space or the low importance of the hyperparameters.

\subsection{Objective Mismatch}
\label{sec:objective_mismatch}
\begin{figure*}[h]
\centering
    \subfigure[Score]{\includegraphics[width=0.43\textwidth]{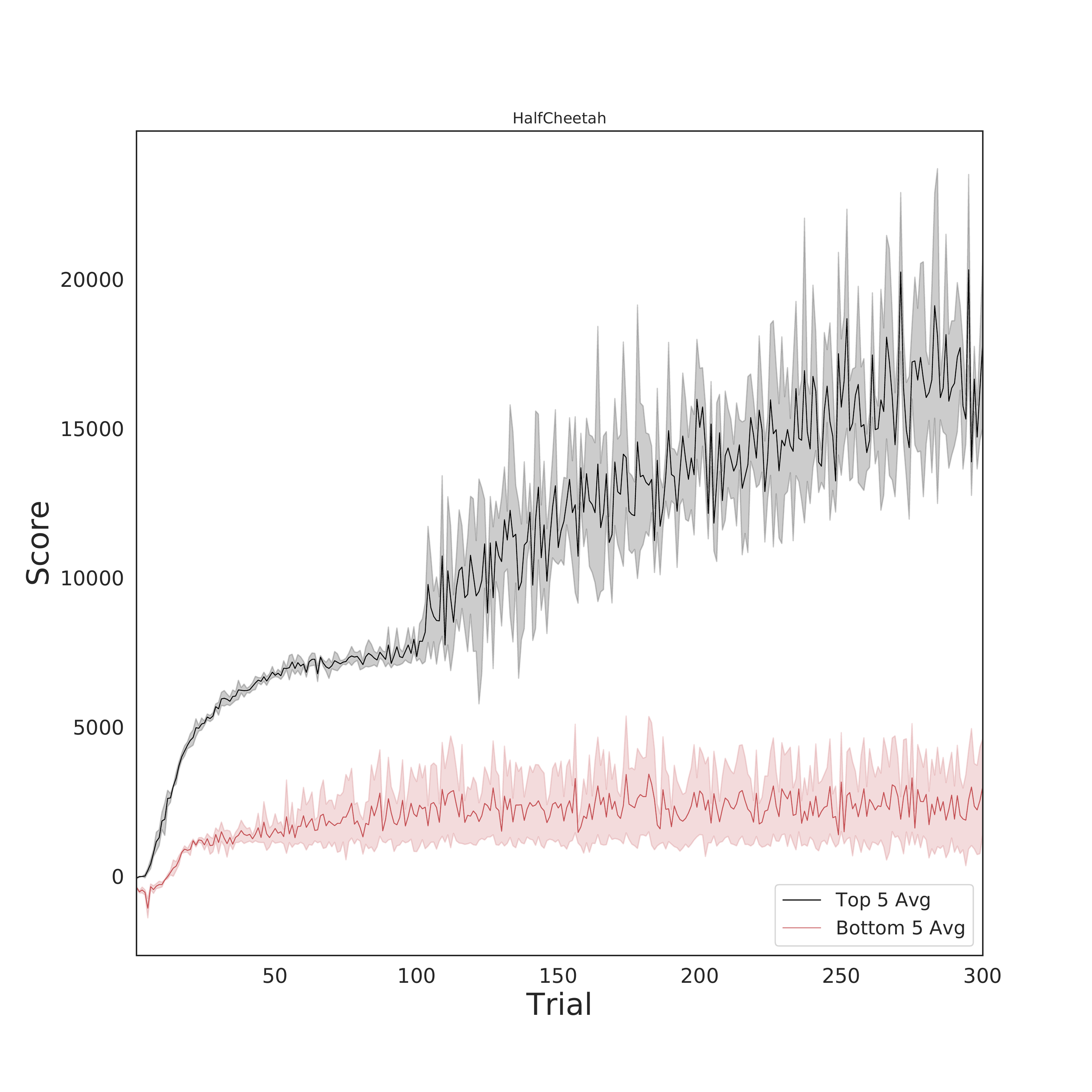}\label{fig:hf_score}}
    \subfigure[Plan Horizon]{\includegraphics[width=0.43\textwidth]{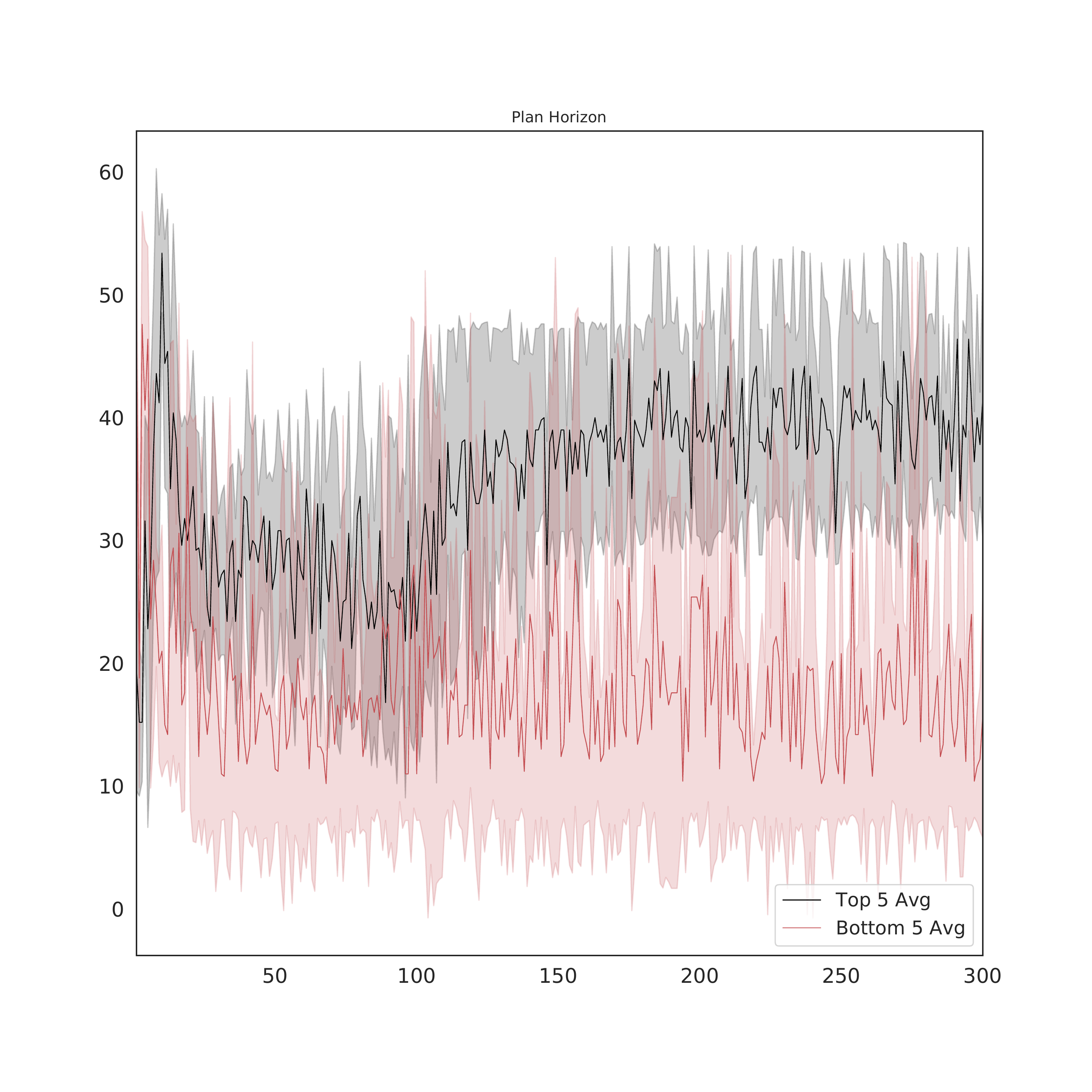}\label{fig:hf_config}}

\caption{Results of using random search to optimize only the plan horizon. 40 models were trained, each with a plan horizon sampled from the configuration space. We plot the average performance and plan horizon of the top $5$ members and the bottom $5$ members over the training.}
\label{fig:hc_plan_horizon}
\end{figure*}

In order to get an in-depth analysis of the relationship between model uncertainties and planning horizon in the context of MBRL and to address the Objective Mismatch problem found by~\citep{lambert2020objective}, we train 40 models on HalfCheetah with different plan horizons. The performance and the learned hyperparameter schedules can be found in Figure~\ref{fig:hc_plan_horizon}. We find that the performance improvement is even larger compared to optimize all \textit{CEM Optimizer} hyperparameters. This may be because \textit{planning horizon} is more important than the other \textit{CEM Optimizer} hyperparameters. Moreover, models with a low plan horizon perform well in the first ~10 trials but do not manage to learn the other gaits which result in higher returns.

\begin{figure*}[h]
    \vspace*{-3mm}
    \centering
    \includegraphics[width=0.4\linewidth]{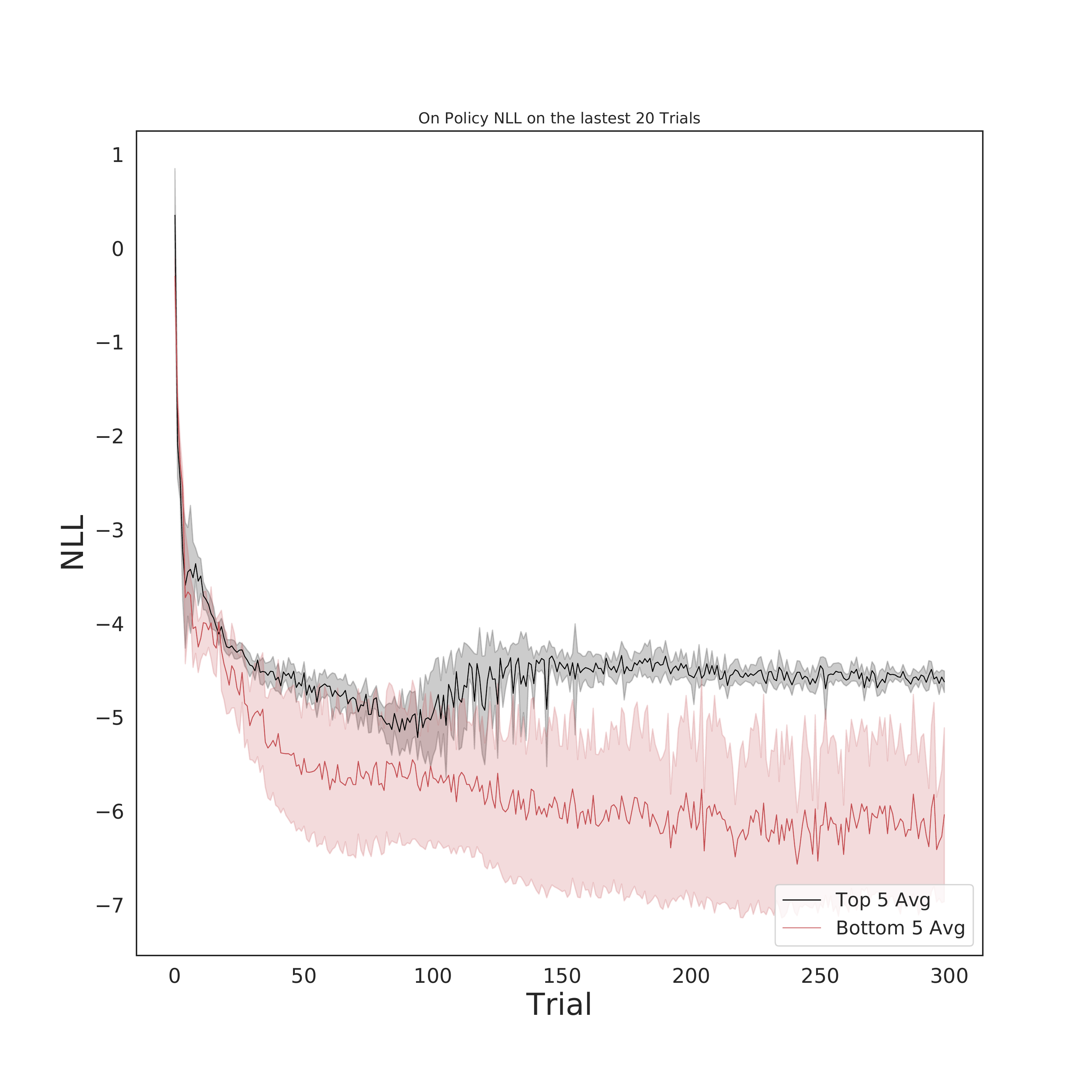}
    \caption{\label{fig:on_policy_nll} NLL of top and bottom members for the objective mismatch experiment. On the \textit{y-axis}, we show the on-policy NLL of each model by evaluating the model on the last 20 trials collected (if not enough trials are available, then we use all trials collected until that time point) and the number of trials on the \textit{x-axis}.}
\end{figure*}

To evaluate the obtained models, we first evaluate them on their own training data, which we refer to as on-policy evaluation.
This provides more evidence on the objective mismatch problem in MBRL. Figure~\ref{fig:on_policy_nll} shows the on-policy negative log likelihood (NLL) of the top/bottom members' models on the trajectories they collected in the last 20 trials. We find that the bottom members have lower NLL compared to the top members. This is because the bottom members only explored the trajectories that gave similarly poor behaviors (in HalfCheetah, the agents kept falling on the ground) and they overfit to the data generated from such behaviors. This, again, highlights the objective mismatch problem in MBRL - lower losses when fitting the model is not the same as the objective of the planning algorithm and, thus, does not translate to better performance.

\begin{figure}[h]
    \centering
    \subfigure[Low return]{\includegraphics[width=0.32\textwidth]{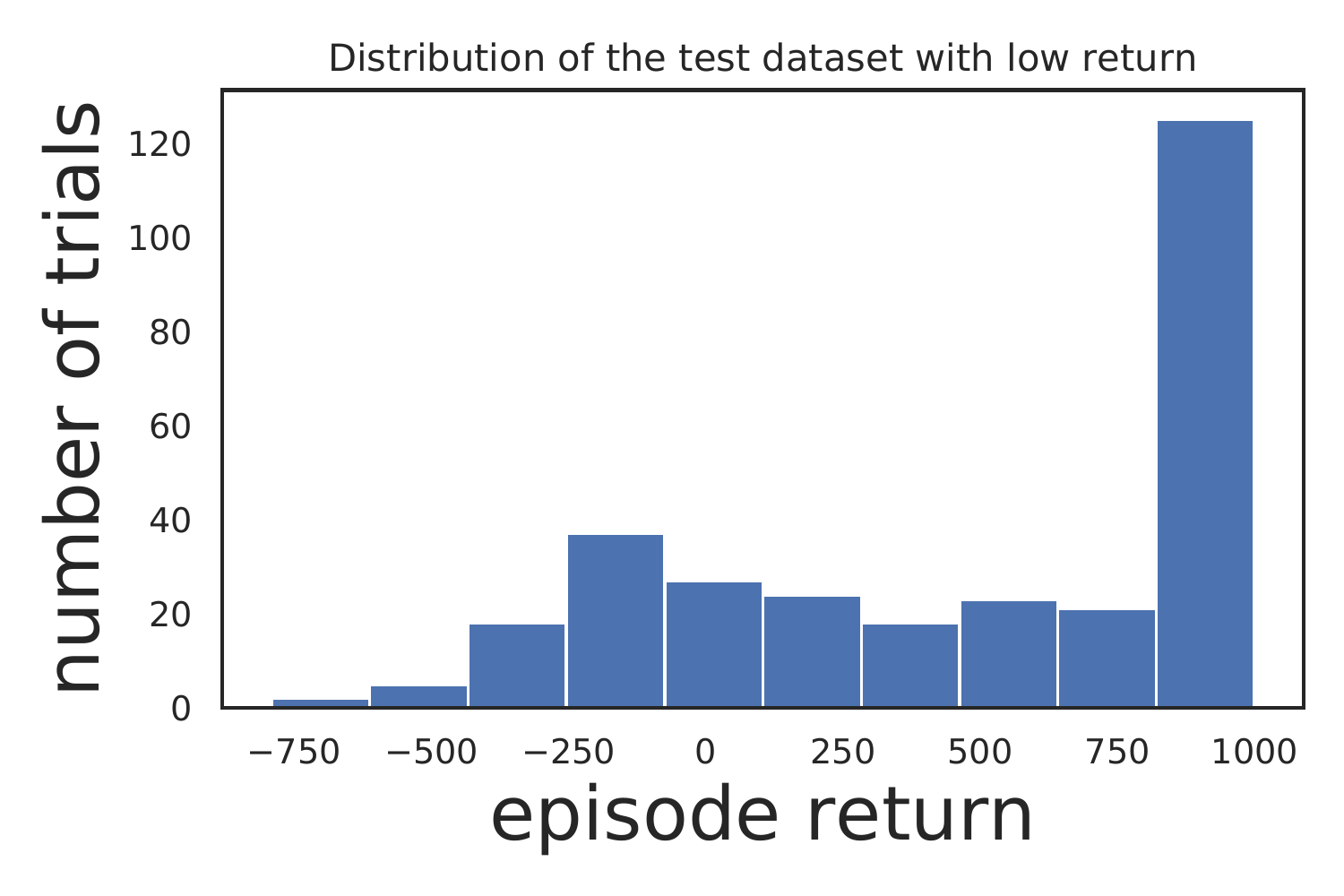}\label{fig:dist_low_ret}}
    \subfigure[Medium return]{\includegraphics[width=0.32\textwidth]{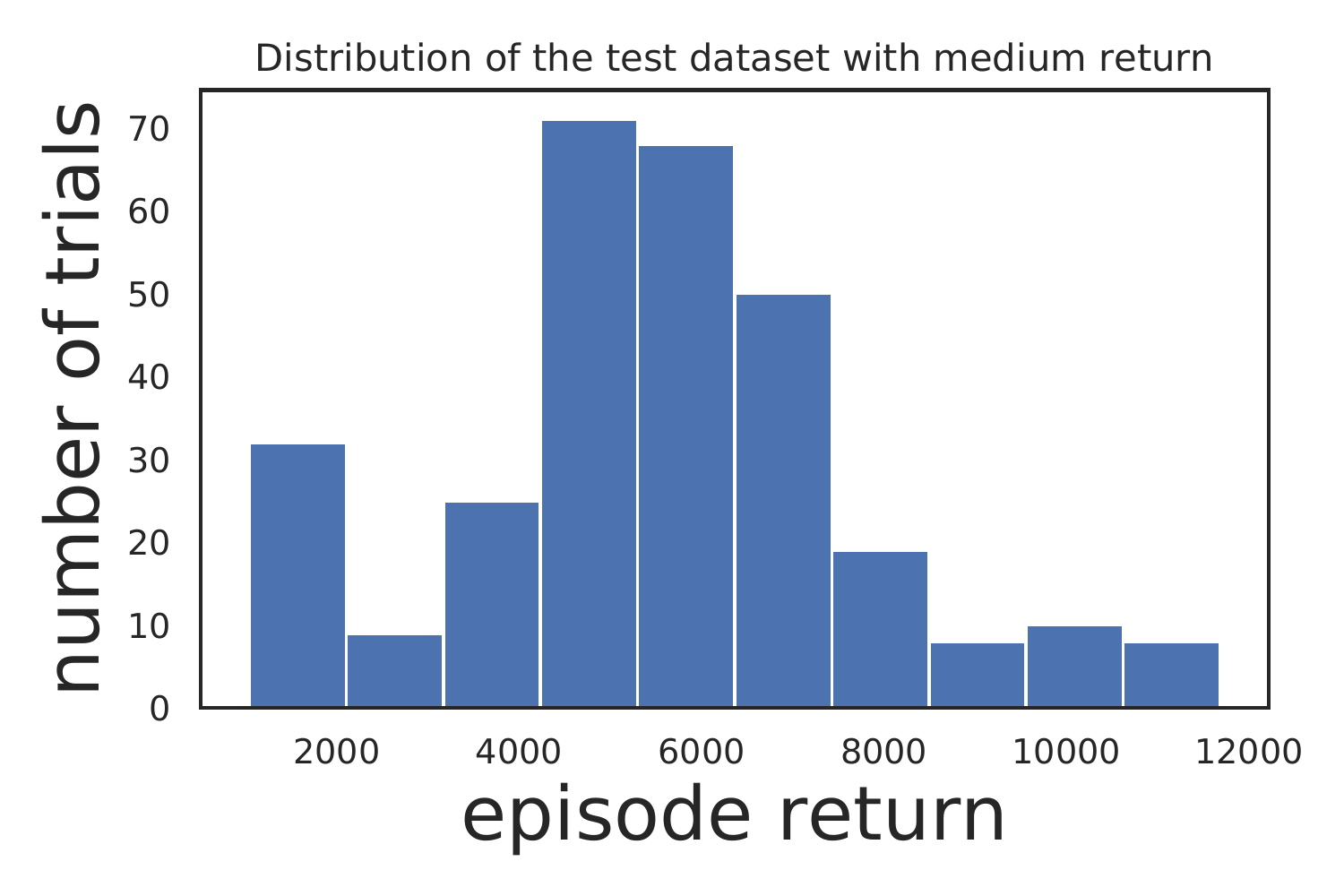}\label{fig:dist_mid_ret}}
    \subfigure[High return]{\includegraphics[width=0.32\textwidth]{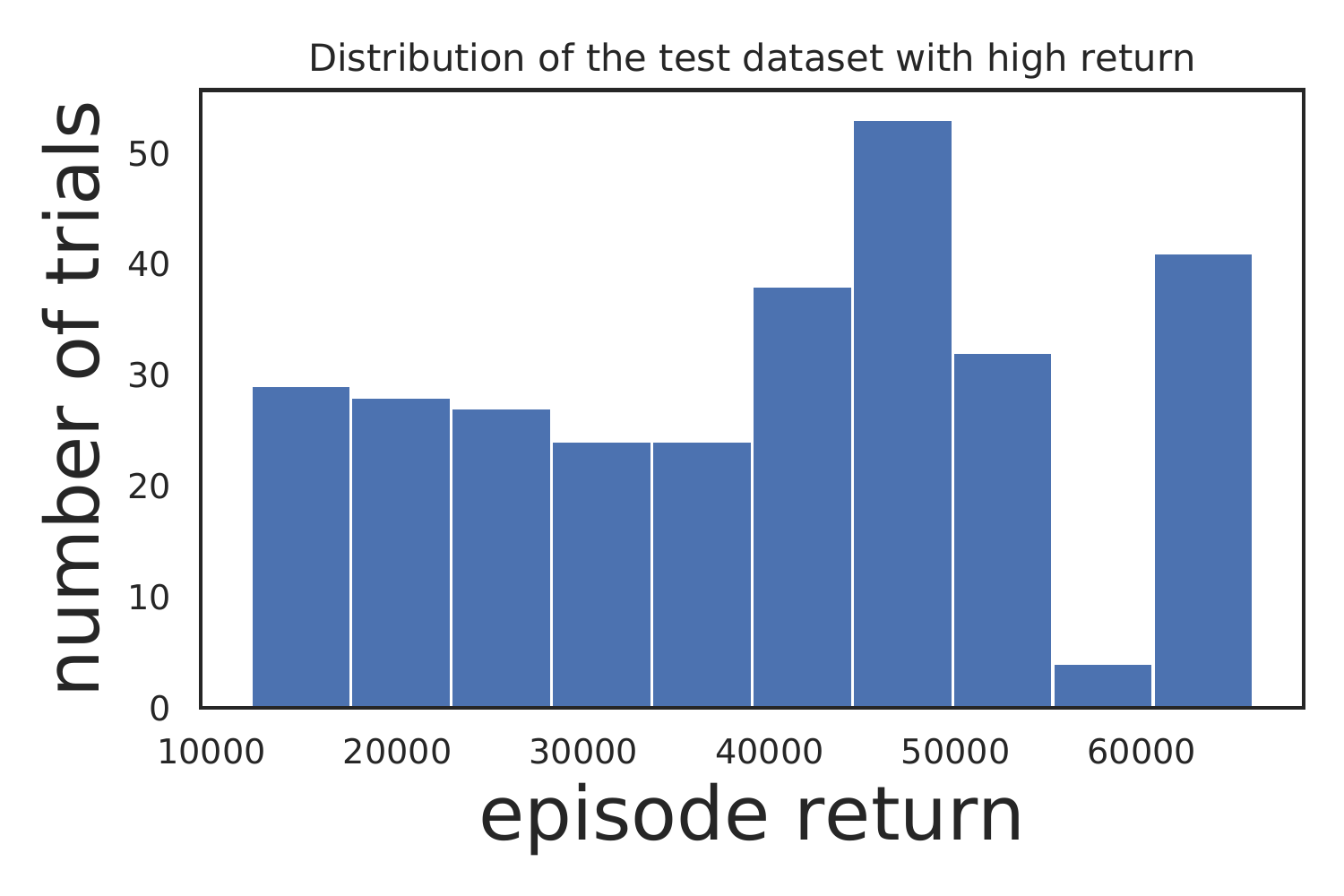}\label{fig:dist_high_ret}} 
    \hfil
    \centering
    \subfigure[Low return]{\includegraphics[width=0.32\textwidth]{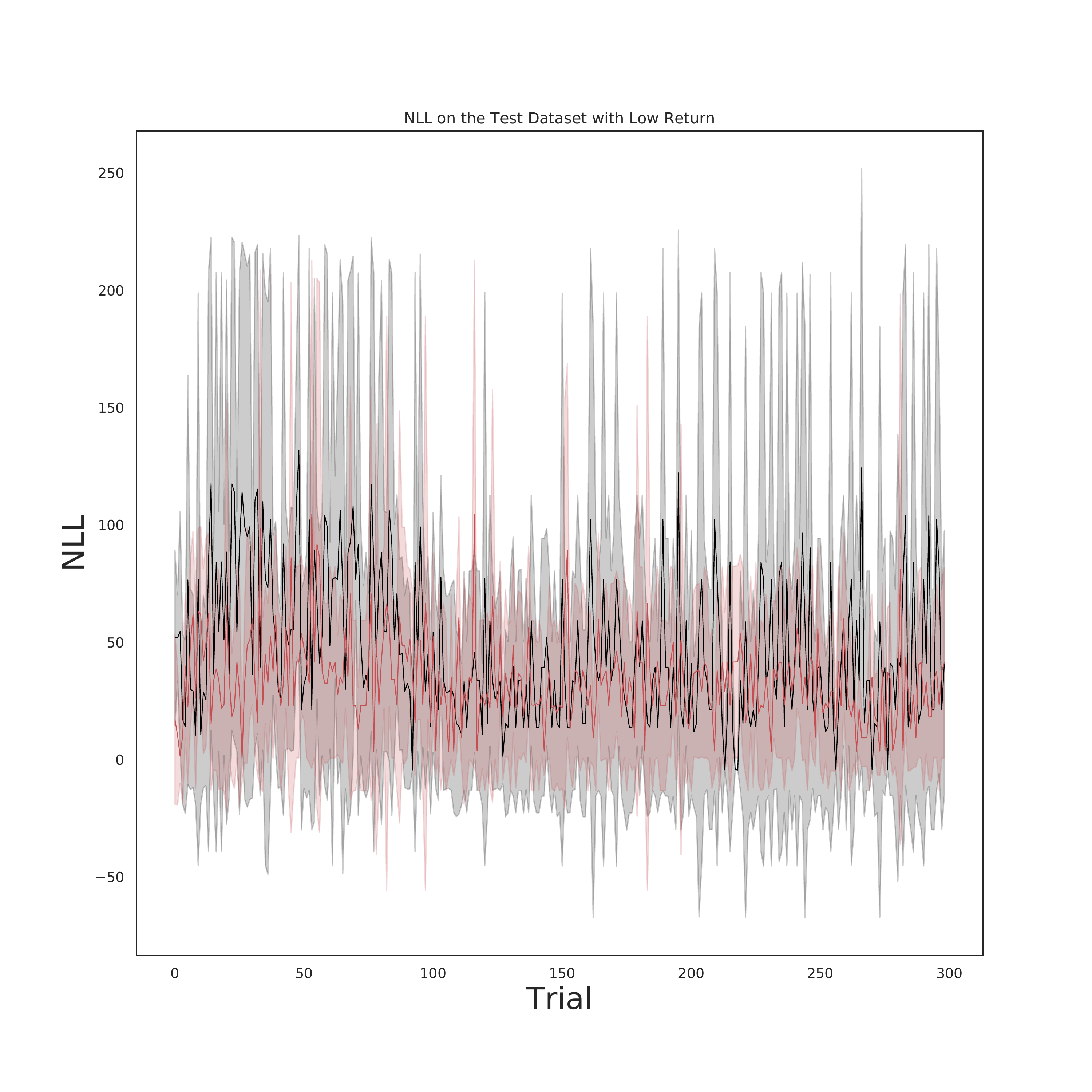}\label{fig:hc_low_ret_nll}} 
    \subfigure[Medium return]{\includegraphics[width=0.32\textwidth]{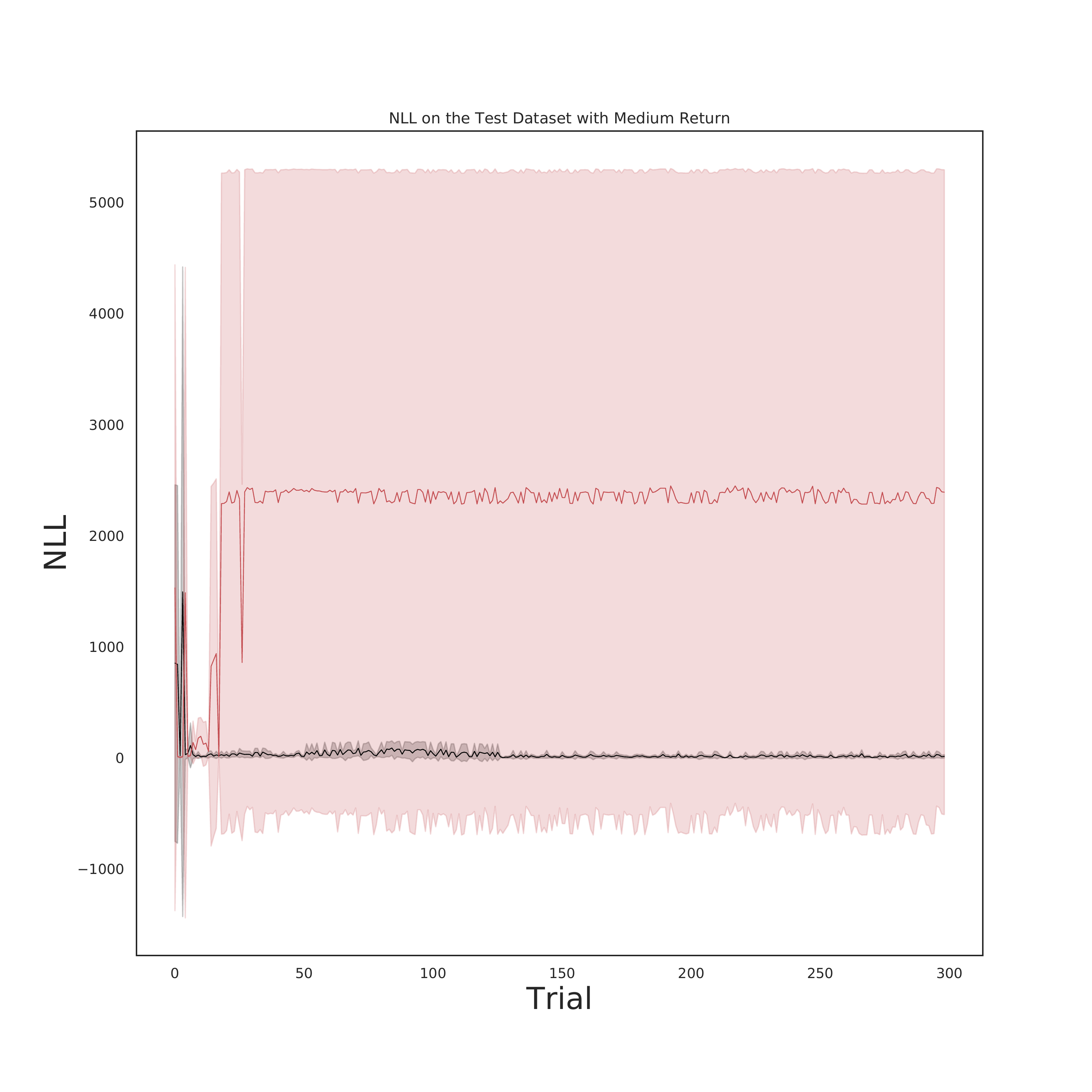}\label{fig:hc_mid_ret_nll}}
    \subfigure[High return]{\includegraphics[width=0.32\textwidth]{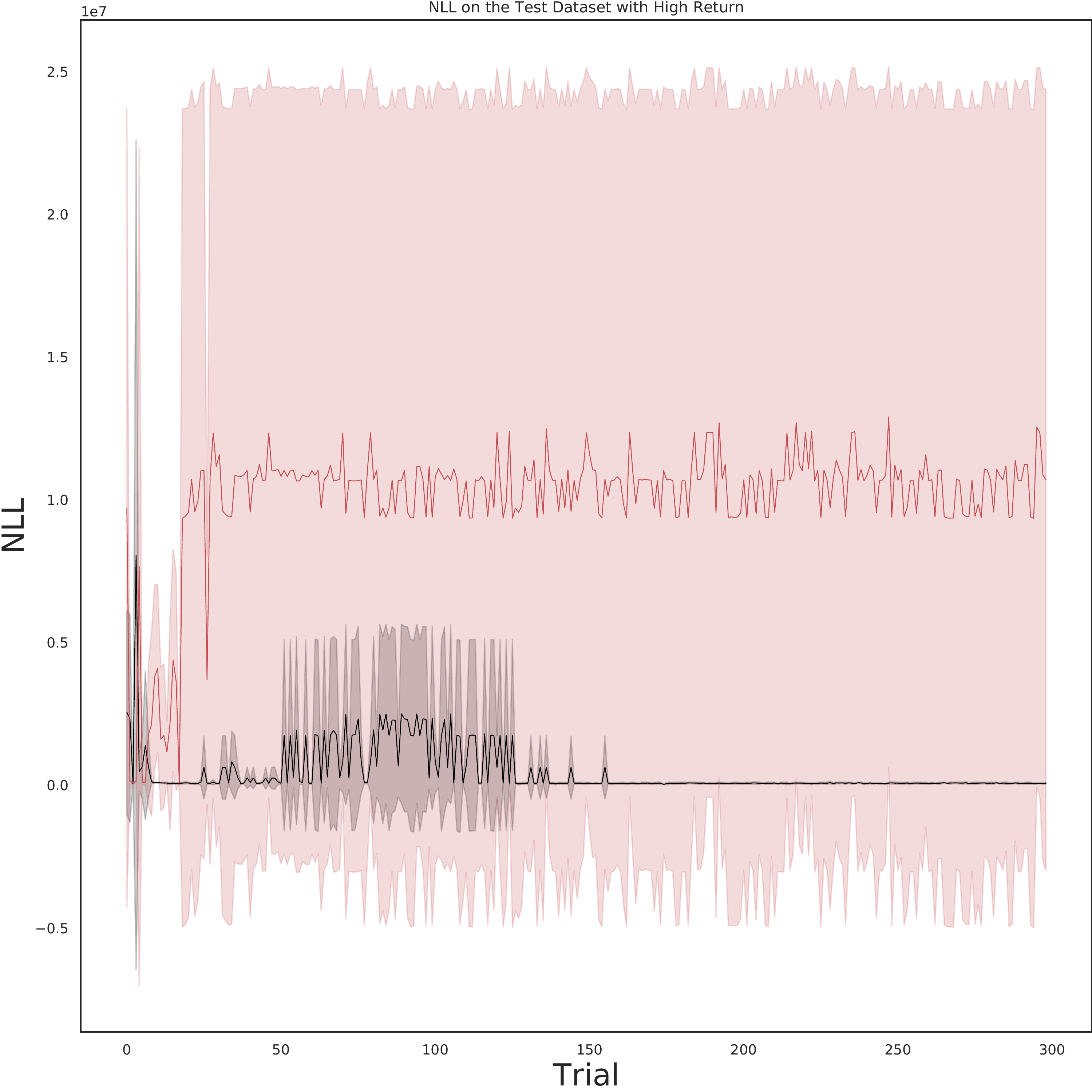}\label{fig:hc_high_ret_nll}} 

    \caption{(Upper row) Reward distribution of three "out-of-distribution" test datasets. Data is generated by different policies and therefore not seen by the models during their training. Each dataset has 300 trajectories. (Lower row) The average NLL of the top/bottom members on the test dataset during training.}
    
    \label{fig:distribution_eval_dataset}
\end{figure}

Secondly, we evaluate them on \textit{unseen} data generated from other policies, which we refer to as off-policy evaluation.
The test dataset consists of trajectories from the other policies. We split the test dataset into 3 subsets, each consisting of 300 trajectories, based on the episodic return, to further analyse the uncertainties of the models in different data regimes. The distribution of the three datasets and evaluation results on these three data regimes can be found in Figure~\ref{fig:distribution_eval_dataset}. From figures~\ref{fig:hc_mid_ret_nll} \& \subref{fig:hc_high_ret_nll}, We see that the models' uncertainties increase dramatically for the poorly performing models which consistently explored only the poorly performing behaviors during the training and did not learn any meaningful strategies. For the well performing models which learned good policies, however, the NLL do not explode as for the poorly performing ones. In fact, we see much lower loss for well-performing models compared to poorly performing models when evaluated off-policy while the opposite was the case for the on-policy evaluation. This supports the conclusion that poorly performing models are overfitting to their training data while well-performing models are able to generalise to unseen data.
Both, top models and bottom models, have similar NLL in the low return data regime. However, models which have a larger plan horizon have lower returns initially (See Figure~\ref{fig:hc_plan_horizon}), which supports our hypothesis that in order to perform long-term planning, models first need to be accurate enough. In words, when models have high uncertainties in an environment, the plan horizon should be relatively low to reduce the uncertainties introduced in the planning algorithms.